\documentclass[runningheads]{llncs}
\usepackage{graphicx}
\usepackage{amsmath,amssymb} 
\usepackage{color}
\usepackage{epsfig}

\usepackage[pagebackref=true,breaklinks=true,letterpaper=true,colorlinks,bookmarks=false]{hyperref}

\usepackage{times}  
\usepackage{helvet} 
\usepackage{courier}  
\usepackage{appendix}

\usepackage{booktabs}
\usepackage{placeins}
\usepackage{float}

\begin{document}
\pagestyle{headings}
\mainmatter

\def\ACCV20SubNumber{560}  

\title{Sketch-to-Art: Synthesizing Stylized Art Images From Sketches} 
\titlerunning{Sketch2art: Synthesizing Stylized Art Images From Sketches}
\authorrunning{Liu et al.}

\author{Bingchen Liu\inst{1,2} \and Kunpeng Song\inst{1,2} \and Yizhe Zhu\inst{2} \and Ahmed Elgammal\inst{1,2}}
\institute{
Playform - Artrendex Inc., USA \\
\email{\{bingchen, kunpeng, elgammal\}@artrendex.com}\\
\url{https://www.playform.io/} \\\and
Department of Computer Science, Rutgers University, USA \\
\email{yizhe.zhu@rutgers.edu}}

\maketitle

\begin{abstract}
We propose a new approach for synthesizing fully detailed art-stylized images from sketches. Given a sketch, with no semantic tagging, and a reference image of a specific style, the model can synthesize meaningful details with colors and textures. Based on the GAN framework, the model consists of three novel modules designed explicitly for better artistic style capturing and generation. To enforce the content faithfulness, we introduce the dual-masked mechanism which directly shapes the feature maps according to sketch. To capture more artistic style aspects, we design feature-map transformation for a better style consistency to the reference image. Finally, an inverse process of instance-normalization disentangles the style and content information and further improves the synthesis quality. Experiments demonstrate a significant qualitative and quantitative boost over baseline models based on previous state-of-the-art techniques, modified for the proposed task (17\% better Frechet Inception distance and 18\% better style classification score). Moreover, the lightweight design of the proposed modules enables the high-quality synthesis at $512 \times 512$ resolution.
\end{abstract}

\begin{figure}
\centering
  \includegraphics[width=0.9\linewidth,height=0.36\linewidth]{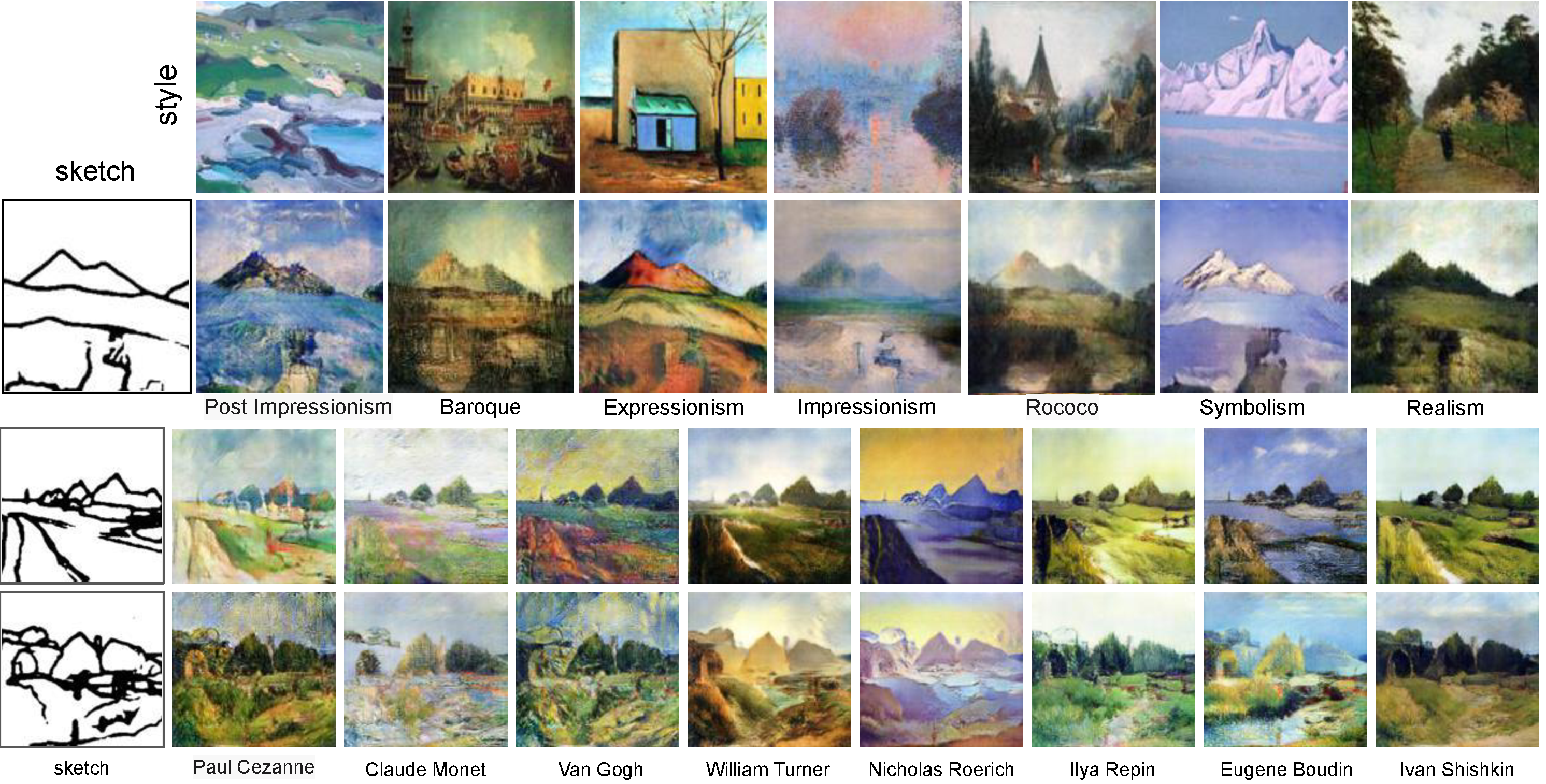}
  \caption{Synthetic images from sketches with different styles. Upper panel:  our approach synthesizes from different styles (art movements). The first row shows the reference images from each style, and the second row shows the generated images. Lower panel:  our model synthesizes from specific artists' styles by taking paintings from the artists as reference.}
  \label{fig:hero1}
\end{figure}

\section{Introduction}
Synthesizing fully colored images from human-drawn sketches is an important problem, with several real-life applications. For example, colorizing sketches following a specified style can significantly reduce repetitive works in story-boarding. Fruitful results have been achieved in applying deep learning to the art literature \cite{elgammal2017can,elgammal2018shape,kim2018finding}. Most research works have focused on synthesizing photo-realistic images \cite{chen2018sketchygan}, or cartoonish images \cite{zhang2017style} from sketches. In this paper, we focus on rendering an image in a specific given artistic style based on a human-drawn sketch as input. The proposed approach is generic, however, what distinguish art images from other types of imagery is the variety of artistic styles that would affect how a sketch should be synthesized into a fully colored and textured image. 

In the history of art, style can refer to an art movement (Renaissance style, Baroque style, Impressionism, etc.), or particular artist style (Cezanne style, Monet style, etc.), or a specific artwork style \cite{schapiro1994theory}. Style encompasses different formal elements of art, such as the color palette, the rendering of the contours (linear or painterly), the depth of the field (recessional or planer), the style of the brush strokes, the light (diffused or light-dark contrast), etc. 

We propose a novel generation task: given an input sketch and a style, defined by a reference image, or an artist name, or a style category, we want to synthesize a fully colored and textured image in that style following the sketch, as shown in Figure~\ref{fig:hero1}. Previous works on synthesizing from sketches do not allow users to specify a style reference~\cite{sangkloy2017scribbler,chen2018sketchygan}. We propose a new model to achieve this task, which takes the input sketch and sample reference image(s) from art-historical corpus, defining the desired style, to generate the results.    

A sketch contains very sparse information, basically the main composition lines. The model has to guess the semantic composition of the scene and synthesize an image with semantic context implied from the training corpus. E.g., given a corpus of landscape art of different styles, the model needs to learn how different parts of the scene correlate with colors and texture given a choice of style. In the proposed approach, no semantic tagging is required for the sketches nor the style images used at both training and generation phases. The model implicitly infers the scene semantic. 

The proposed approach is at the intersection between two different generation tasks: Sketch-to-image synthesis and Style transfer. Sketch-to-image synthesis focuses on rendering a colored image based on a sketch, where recent approaches focused on training deep learning models on a corpus of data, e.g., \cite{sangkloy2017scribbler}. However, these approaches do not control the output based on a given style. 

More importantly, in our case the reference image should only define the style not the content details.
For example, the model should infer that certain plain area of the sketch is sky, trees, mountains, or grass; and infer details of these regions differently given the style image, which might not have the same semantic regions all together (see examples in Figure~\ref{fig:hero1}). 

\begin{figure}
\centering
\includegraphics[width=0.9\linewidth,height=0.42\linewidth]{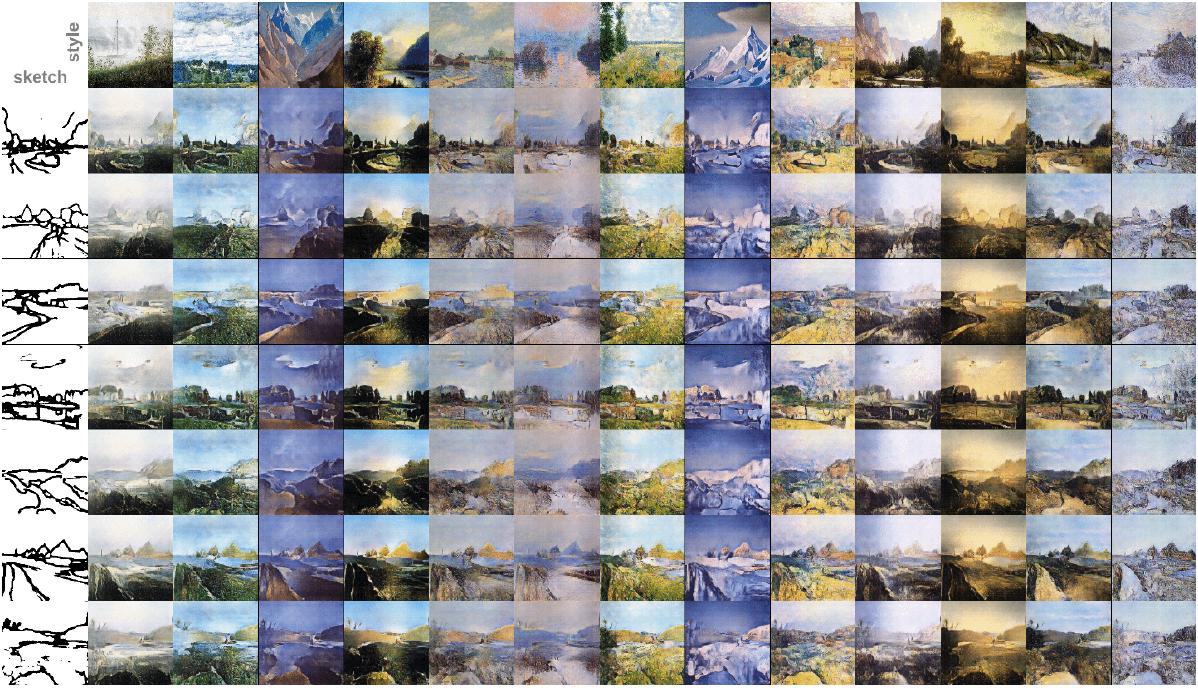}
  \caption{Synthesized samples in $512\times512$ resolution. The first column are the input sketches hand-drawn by human, and the first row are the style reference images.}
  \label{fig:hero2}
\end{figure}

On the other hand, style transfer focuses on capturing the style of an (or many) image and transfer it to a content image \cite{gatys2015neural}. However, we show that such models are not applicable to synthesize stylized images from sketches because of the lack of content in the sketch, i.e., the approach need to both transfer style from the reference image and infer content based on the training corpus.

The proposed model has three novel components, which constitute the technical contributions of this paper, based on a GAN \cite{goodfellow2014generative} infrastructure,:

  \noindent{\bf Dual Mask Injection.} A simple trainable layer that directly imposes sketch constraints on the feature maps, to increase content faithfulness.
  
  \noindent{\bf Feature Map Transfer.} An adaptive layer that applies a novel transformation on the style image's feature map, extracting only the style information without the interference of the style images' content. 
  
  \noindent{\bf Instance De-Normalization.} A reverse procedure of Instance Norm \cite{ulyanov2016instance} on the discriminator to effectively disentangle the style and content information.

\section{Related Work}
\noindent\textbf{Image-to-Image Translation:} I2I aims to learn a transformation between two different domains of images. The application scenario is broad, including object transfiguration, season transfer, and photo enhancement. With the generative power of GAN \cite{goodfellow2014generative}, fruitful advances have been achieved in the I2I area. Pix2pix \cite{isola2017image} established a common framework to do the one-to-one mapping for paired images using conditional GAN. Then a series of unsupervised methods such as CycleGAN \cite{zhu2017unpaired,liu2017unsupervised,choi2018stargan,yi2017dualgan} were proposed when paired data is not available. Furthermore, multi-modal I2I methods are introduced to simulate the real-world many-to-many mapping between image domains such as MUNIT and BicycleGAN \cite{huang2018multimodal,zhu2017toward,almahairi2018augmented,lee2018diverse,liu2020time,zhu2020s3vae}. However, those I2I methods can not generate satisfying images from the coarse sketch's domain, nor can they adequately reproduce the styles from the reference image, as will be shown in the experiments. 

\noindent\textbf{Neural Style Transfer:} NST transfers textures and color palette from one image to another. In \cite{gatys2015neural}, the problem was formulated by transforming the statistics of multi-level feature maps in CNN layers in the form of a Gram Matrix. The follow-up works of NST have multiple directions, such as accelerating the transfer speed by training feed-forward networks \cite{johnson2016perceptual,zhang2017multistyle}, and capturing multiple styles simultaneously with adaptive instance normalization \cite{huang2017arbitrary,ulyanov2016instance}. However, little attention has been paid on optimizing towards artistic style transfer from a sketch, nor on improving the style transfer quality regarding the fine-grained artistic patterns among the various NST methods \cite{jing2017neural}. 

\noindent\textbf{Sketch-to-image Synthesis:} Our task can be viewed from both the I2I and the NST perspectives, while it has its unique challenges. Firstly, unlike datasets such as horse-zebra or map-satellite imagery, the sketch-to-painting dataset is heavily unbalanced and one-to-many. The information in the sketch domain is ambiguous and sparse with only few lines, while in the painting's domain is rich and diverse across all artistic styles. Secondly, ``style transfer'' approaches focus mainly on color palettes and ``oil painting like" textures but pays limited attention to other important artistic attributes such as linear vs. painterly contours, texture boldness, and brush stroke styles. Thirdly, it is much harder for NST methods to be semantically meaningful, as the optimization procedure of NST is only between few images. On the contrary, with a sketch-to-image model trained on a large corpus of images, learning semantics is made possible (the model can find the common coloring on different shapes and different locations across the images) and thus make the synthesized images semantically meaningful. 

To our knowledge, there are few prior works close to our task. AutoPainter \cite{liu2017auto} propose to do sketch-to-image synthesis using conditional GANs, but their model is designed towards cartoon images. ScribblerGAN \cite{sangkloy2017scribbler} achieves user-guided sketch colorization but requires the user to provide localized color scribbles. SketchyGAN \cite{chen2018sketchygan} is the state-of-the-art approach for multi-modal sketch-to-image synthesis, focusing on photo-realistic rendering of images conditioned on object categories. Furthermore, \cite{zhang2017style,zhang2018two,park2019semantic} accomplish the conditioned colorization with a reference image, but they only optimize towards datasets that lack style and content variances. None of those works is specifically geared towards artistic images with the concern of capturing the significant style variances. In contrast, Figure~\ref{fig:hero1} and Figure~\ref{fig:hero2} demonstrate the ability of our model to capture the essential style patterns and generate diversely styled images.

\section{Methods}
In this section, we provide an overview of our model and introduce the detailed training schema. Then three dedicated components will be described.

\begin{figure}
\centering
  \includegraphics[width=0.85\linewidth,height=0.26\linewidth]{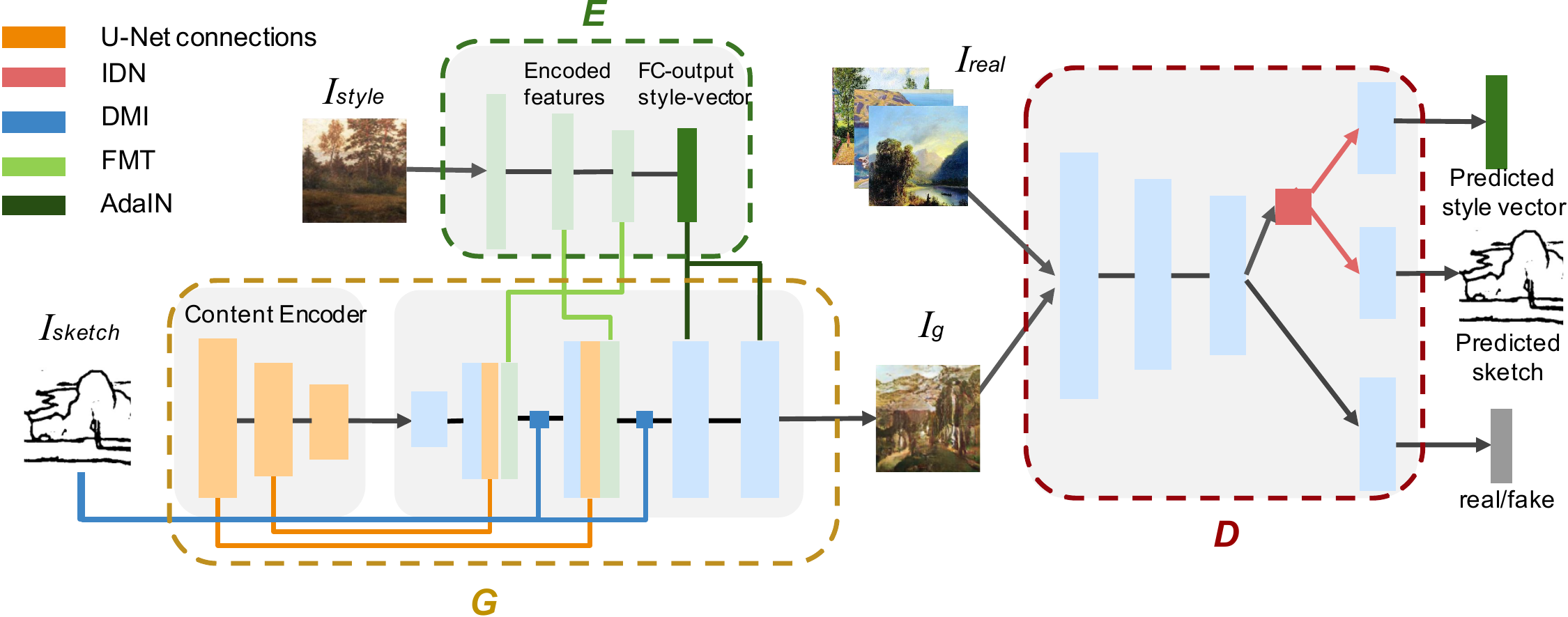}
  \caption{Overview of the model structure. $G$ adopts a U-Net structure \cite{ronneberger2015u}. It takes the features of the reference image $I_{style}$ from $E$, runs style-conditioned image synthesis on the input sketch image $I_{sketch}$, and outputs $I_g$. $D$ takes as input an image (alternatively sampled from real and generated images) and gives three outputs: a predicted style vector, a predicted sketch, and a real/fake signal of the image}
  \label{fig:g}
\end{figure}

As shown in Figure~\ref{fig:g}, our model consists of three parts: a generator $G$, a discriminator $D$, and a separately pre-trained feature-extractor $E$. Either an unsupervised Auto-Encoder or a supervised DNN classifier, such as VGG \cite{simonyan2014very} can work as the feature-extractor. During the training of $G$ and $D$, $E$ is fixed and provides multi-level feature-maps and a style vector of the reference image $I_{style}$. The feature-maps serve as inputs for $G$, and the style vector serves as the ground truth for $D$. We train $G$ and $D$ under the Pix2pix~\cite{isola2017image} schema. In the synthesis phase, the input style is not limited to one reference image. We can always let $E$ extract style features from multiple images and combine them in various ways (averaging or weighted averaging) before feeding them to $G$. Apart from the following three modules, our model also includes a newly designed attentional residual block and a patch-level image gradient matching loss, please refer to the appendix for more details.

\subsection{Reinforcing Content: Dual Mask Injection}
In our case, the content comes in the form of a sketch, which only provides sparse compositional constraints. It is not desirable to transfer composition elements from
the style image or training corpus into empty areas of the sketch that should imply textured areas. Typically when training a generator to provide images with diverse style patterns, the model tends to lose faithfulness to the content, which results in missing or excrescent shapes, objects, and ambiguous contours (especially common in NST methods). To strengthen the content faithfulness, we introduce Dual-Mask Injection layer (DMI), which directly imposes the sketch information on the intermediate feature-maps during the forward-pass in $G$. 

Given the features of a style image, in the form of the conv-layer activation $f\in{\rm I\!R}^{C \times H \times W}$,
we down sample the binary input sketch to the same size of $f$  and use it as a feature mask, denoted as $M_s\in{\rm [0,1]}^{H\times W}$. The proposed DMI layer will first filter out a \textit{contour-area feature} $f_c$ and a \textit{plain-area feature} $f_p$ by:
\begin{align}\label{eq:dmi_1}
f_c = M_s \times f, \quad
f_p = (1 - M_s) \times f,
\end{align}
where $\times$ is element-wise product. For $f_c$ and $f_p$, the DMI layer has two sets of trainable weights and biases, $w_c, b_c, w_p, b_p \in {\rm I\!R}^{ C \times 1 \times 1}$, that serve for a \textbf{value relocation} purpose, to differentiate the features around edge and plain area:
\begin{align}\label{eq:dmi_2}
f_c' = w_c \times f_c + b_c, \quad
f_p' = w_p \times f_p + b_p
\end{align}
Finally, the output feature maps of DMI will be $f' = f_c' + f_p'$.

\begin{figure}
\centering
  \includegraphics[width=0.66\linewidth,height=0.22\linewidth]{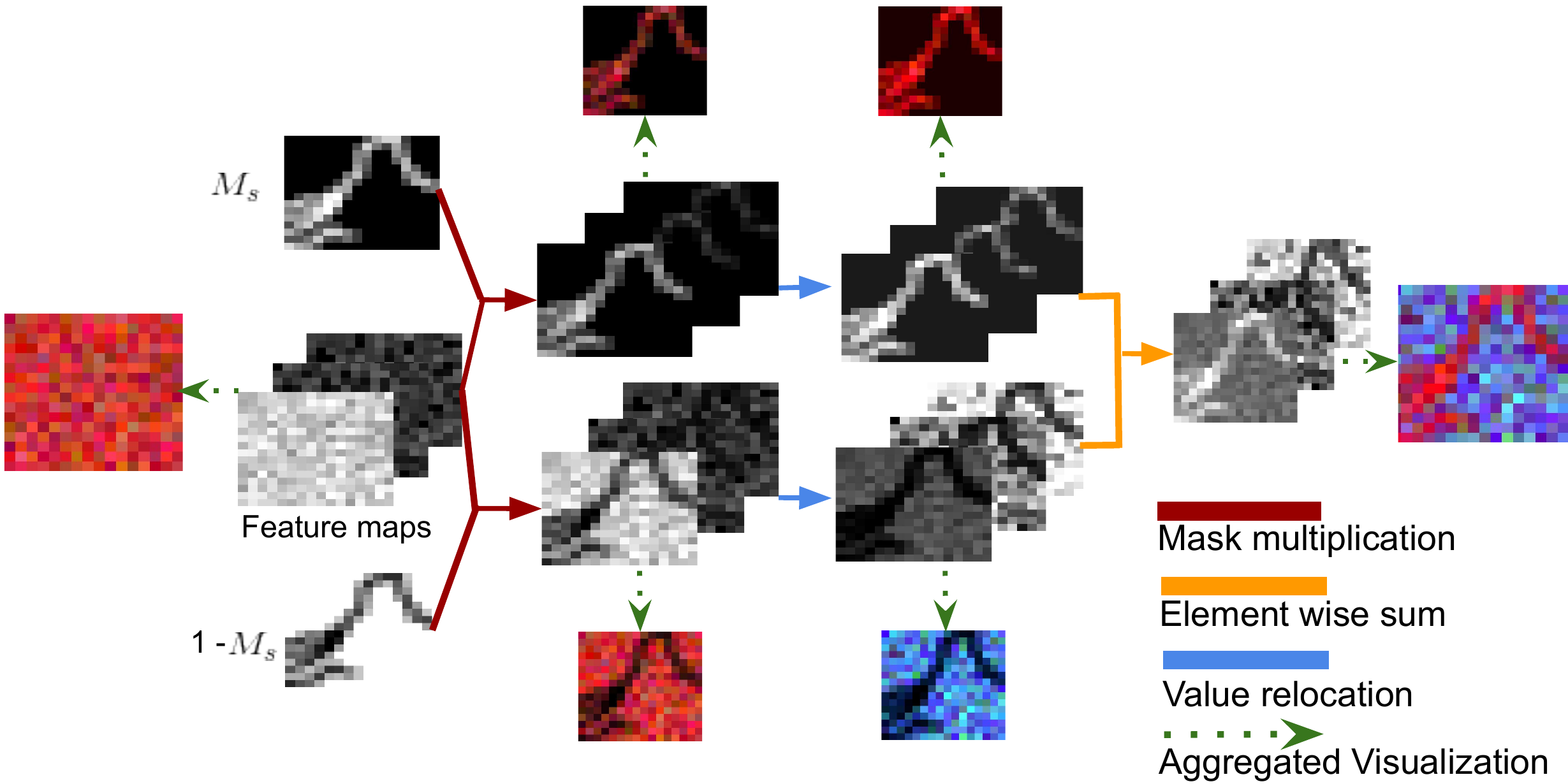}
  \caption{Forward flow of Dual-Mask Injection layer, we aggregate the first three channels in feature-map as an RGB image for visualization purpose}
  \label{fig:DMI}
\end{figure}

A real-time forward flow of the DMI layer is shown in Figure~\ref{fig:DMI}. Notice that, when $w=1$ and $b=0$, the output feature will be the same as the input, and we set the weights and bias along the channel dimension so that the model can learn to impose the sketch on the feature-maps at different degrees on different channels. By imposing the sketch directly to the feature-maps, DMI layer ensures that the generated images have correct and clear contours and compositions. While DMI serves the same purpose as the masked residual layer (MRU) in SketchyGAN \cite{chen2018sketchygan}, it comes with almost zero extra computing cost, where MRU requires three more convolution layers per unit. In our experiments, our model is two times faster in training compared with SketchyGAN, while yields better results on art dataset. Moreover, the lightweight property of DMI enables our model to achieve great performance on $512 \times 512$ resolution, while SketchyGAN was studied only on $128 \times 128$ resolution.

\subsection{Disentangling Style and Content by Instance De-Normalization}
To better guide $G$, $D$ is trained to adequately predict an style latent representation of the input image as well as the content sketch that should match with the input image. Ideally, a well-disentangled discriminator can learn a style representation without the interference of the content information, and retrieve the content sketch regardless of the styles.

Several works have been done in specifically disentangling style and content \cite{kazemi2019style,karras2019style}, but they only work around the generator side, using AdaIN or Gram-matrix to separate the factors. To train $D$ to effectively disentangle, we propose the Instance De-Normalization (IDN) layer. IDN takes the feature-map of an image as input, then reverses the process of Instance-Normalization \cite{huang2017arbitrary,ulyanov2016instance} to produces a style vector and a content feature map. In the training phrase of our model, IDN helps $D$ learn to predict accurate style-vectors and contents of the ground truth images, therefore, helps $G$ to synthesis better. 

In AdaIN or IN, a stylized feature-map is calculated by:
\begin{align}\label{eq:adain}
f_{styled}= \sigma_{style} \times \frac{ (f - \mu(f)) }{ \sigma(f) } + \mu_{style},
\end{align}
where $\sigma(\cdot)$ and $\mu(\cdot)$ calculate the variance and mean of the feature map $f$ respectively. It assumes that while the original feature map $f\in{\rm I\!R}^{C \times H \times W}$ contains the content information, some external $\mu_{style}$ and $\sigma_{style}$ can be collaborated with $f$ to produce a stylized feature map. The resulted feature map possesses the style information while also preserves the original content information.

Here, we reverse the process for a stylized feature map $f_{style}'$ by: first predict $\mu_{style}$ and $\sigma_{style}$, ($\mu_{style}, \sigma_{style}$)$\in{\rm I\!R}^{C \times 1 \times 1}$ from $f_{style}'$, then separate them out from $f_{style}'$ to get $f_{content}$ which carries the style-invariant content information. Formally, the IDN process is:
\begin{align}\label{eq:IDN}
\mu_{style}' = &Conv(f_{style}'), \quad
\sigma_{style}' = \sigma(f_{style}' - \mu_{style}'), \\
&f_{content} = \frac{(f_{style}' - \mu_{style}')}{\sigma_{style}'},
\end{align}
where $Conv(\cdot)$ is 2 conv layers. Note that unlike in AdaIN where $\mu_{style}$ can be directly computed from the known style feature $f_{style}$, in IDN the ground truth style feature is unknown (we don't have $f_{style}$), thus we should not naively compute the mean of $f_{style}'$. Therefore, we use conv layers to actively predict the style information, and will reshape the output into a vector as $\mu_{style}'$. Finally, we concatenate $\mu_{style}'$ and $\sigma_{style}'$ to predict the style-vector with MLP, and use conv layers on $f_{content}$ to predict the sketch. The whole IDN process can be trained end-to-end with the target style-vector and target-sketch. Unlike other disentangling methods, we separate the style and content from a structural perspective that is straightforward while maintaining effectiveness.

\subsection{Reinforcing Style: Feature Map Transformation}
To approach the conditional image generation task, previous methods such as MUNIT~\cite{huang2018multimodal} and BicycleGAN~\cite{zhu2017toward} use a low-dimensional latent vector of $I_{style}$ extracted from some feature extractors $E$ as the conditioning factor. However, we argue that such vector representation carries limited information in terms of style details. Therefore, it is more effective to directly use feature-maps as a conditional factor and information supplier. 

\begin{figure}
\centering
  \includegraphics[width=0.8\linewidth,height=0.28\linewidth]{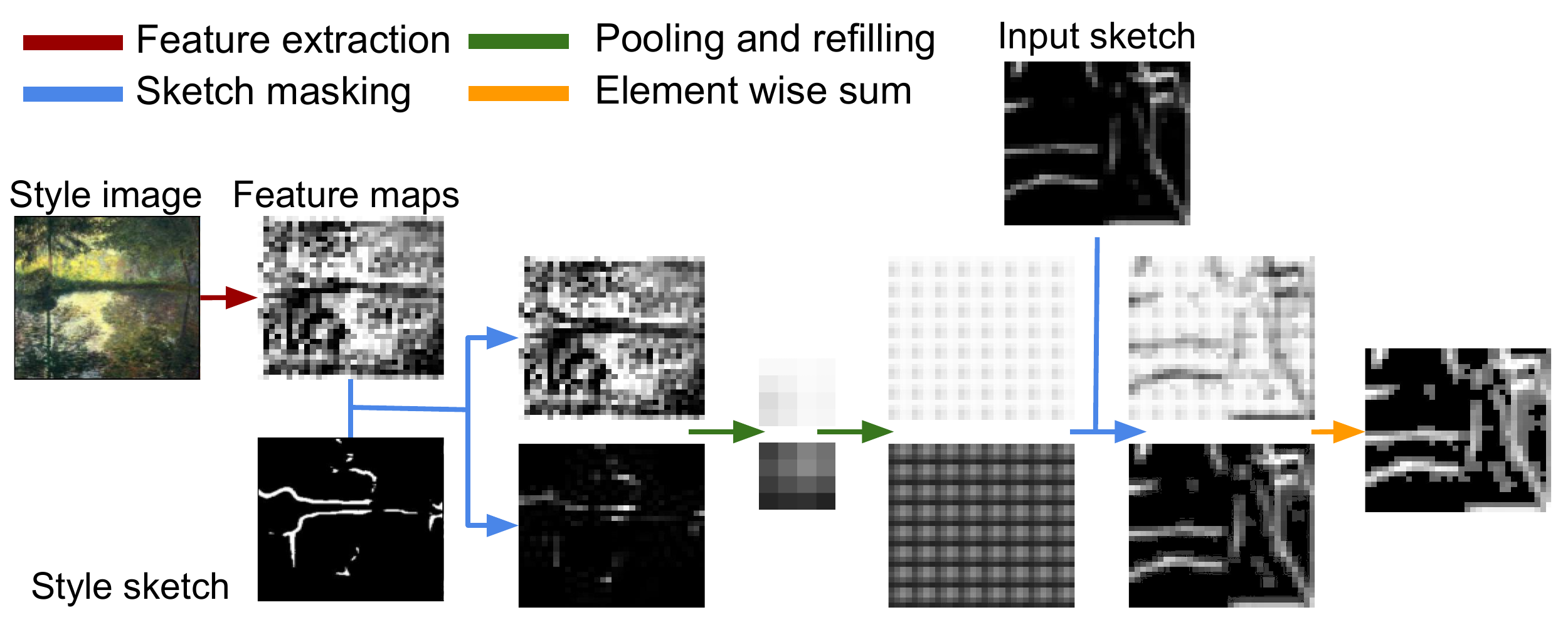}
  \caption{Process flow of Feature-Map Transformation}
  \label{fig:FMT}
\end{figure}

Nevertheless, the image feature-maps in CNN usually carry both the style information and strong content information. Such content information can be problematic and is undesired. For instance, if the style image is a house while the input sketch implies a lake, we do not want any shape or illusion of a house within the lake region in the synthesized image. To get rid of the content information, while keeping the richness of the style features, we propose the Feature Map Transformation (FMT). FMT takes the input sketch $I_{sketch}$, the sketch of the style image $I_{style-sketch}$, and the feature map of the style image $f_{style}$ as inputs, and produces transformed feature-maps $f_t$. $f_t$ only preserves the desired style features of the style image and discards its content structure. Note that $I_{style-sketch}$ is extracted using simple edge detection methods and $f_{style}$ comes from the feature-extractor $E$, that are both easy to achieve.

The proposed FMT is a fixed formula without parameter-training. The procedure is illustrated in Figure~\ref{fig:FMT} with five steps. In step 1, we use $I_{style-sketch}$ as a mask to filter $f_{style}$ and get two sets of features, i.e., $f_{style}^c$ that only have the features around the contours and  $f_{style}^p$ with features on plain areas. In step 2, we apply a series of max-pooling and average-pooling to this filtered yet sparse feature values to extract a $4\times4$ feature-map for each part, namely ${f_{style}^{c'}}$ and ${f_{style}^{p'}}$. In step 3, we repeatedly fill the $4\times4$ ${f_{style}^{c'}}$ and ${f_{style}^{p'}}$ into a $f_t^c$ and a $f_t^p$ with the same size of $f_{style}$. In step 4, we use $I_{sketch}$ as a mask in the same manner to filter these two feature maps, and get ${f_t^c}'$ and ${f_t^p}'$ that have the features of $I_{style}$ but in the shape of $I_{sketch}$. Finally in step 5, we add the results to get $f_t = {f_t^c}' + {f_t^p}'$ as the output of the FMT layer. We then concatenate $f_t$ to its corresponding feature-maps in $G$ for the synthesis process. 

The pooling operation collects the most distinguishable feature values along spatial channel in $f_{style}$, then the repeat-filling operation expands the collected global statistics, finally the masking operation makes sure the transformed feature-map will not introduce any undesired content information of the style image. FMT provides accurate guidance to the generation of fine-grained style patterns in a straightforward manner, and unlike AdaIN \cite{huang2017arbitrary} and Gram-matrix \cite{gatys2015neural} which require higher-order statistics. In practice, we apply FMT from the $16\times16$ to $64\times64$ feature maps. FMT contains two max-pooling layers and one avg-pooling layer. These layers have $5\times5$ kernel and stride of 3, which give us a reasonable receptive field to get the highlighted style features. We first use max-pooling to get the most outstanding feature values as the style information, then use the average-pooling to summarize the values into a “mean feature” along the spatial dimensions. The average-pooling can help smooth out some local bias (peak values) and get a more generalized style representation.

\subsection{Objective Functions}
Besides the original GAN loss, auxiliary losses are adopted in our model. Specifically, when the input is a real image, we train $D$ with a \textit{style loss} and a \textit{content loss}, which minimize the \textit{MSE} of the predicted style vector and sketch with the ground-truth ones respectively. Meanwhile, $G$ aims to deceive $D$ to predict the same style vector as the one extracted from $I_{style}$ and output the same sketch as $I_{sketch}$. These auxiliary losses strengthen $D$'s ability to understand the input images and let $G$ generate more accurate style patterns while ensuring the content faithfulness. During training, we have two types of input for the model, one is the paired data, where the $I_{sketch}$ and $I_{style}$ are from the same image, and the other is randomly matched data, where $I_{sketch}$ and $I_{style}$ are not paired. We also have a reconstruction \textit{MSE} loss on the paired data.

Formally, $D$ gives three outputs: $S(I)$, $C(I)$, $P(I)$, where $S(I)$ is the predicted style vector of an image $I$, $C(I)$ is the predicted sketch, and $P(I)$ is the probability of $I$ being a real image. Thus, the loss functions for $D$ and $G$ are:

\begin{align}
\label{eq:loss_dg}
\mathcal{L}(D) = &\mathop{\mathbb{E}}[\log(P(I_{real}))] + \mathop{\mathbb{E}}[\log(1 - P(G(I_{sketch}, I_{style})))] \nonumber \\
&+MSE(S(I_{real}), E(I_{real})) + 
MSE(C(I_{real}), I_{real-sketch}),\\
\mathcal{L}(G) = &\mathop{\mathbb{E}}[\log(P(G(I_{sketch}, I_{style})))]+MSE(C(G(I_{sketch}, I_{style})), I_{sketch}) \nonumber \\
&+MSE(S(G(I_{sketch}, I_{style})), E(I_{style})).
\end{align}
and the extra loss for G: 
$MSE(G(I_{sketch}, I_{style}), I_{style})$ is applied when the inputs are paired. $I_{real}$ is randomly sampled real data and $I_{real-sketch}$ is its corresponding sketch, and $I_{sketch}$ and $I_{style}$ are randomly sampled sketches and referential style images as the input for $G$.

\section{Experiments}

We first show comparisons between our model and baseline methods, and then present the ablation studies. The code to reproduce all the experiments with detailed training configurations are included in the supplementary materials, along with a video demonstrating our model in real time. A website powered by the proposed model is available online at: https://www.playform.io, where people can synthesize $512\times512$ images with their free-hand drawn sketches.

\noindent\textbf{Dataset:} Our dataset is collected from Wikiart \cite{wikiart} and consists of  10k images with 55 artistic styles (e.g., impressionism, realism, etc.). We follow the sketch creation method described by \cite{chen2018sketchygan} to get the paired sketch for each painting. We split the images into training and testing set with a ratio of $9:1$. All the comparisons shown in this section were conducted on the testing set, where both the sketches and the art images were unseen to the models. 

\noindent\textbf{Metrics: } We use \textbf{FID} \cite{heusel2017gans} and a \textbf{classification accuracy} for the quantitative comparisons. FID is a popular image generation metric that provides a perceptual similarity between two sets of images. For our task, we generate $I_g$ from all the testing sketches using the same $I_{style}$, and compute the FID between $I_g$ and the images from the same style. We repeat this process for all style images. We further employ a more direct metric to compute the style classification accuracy of $I_g$, which leverages a VGG model pre-trained on ImageNet \cite{russakovsky2015imagenet} and fine-tuned on art for style classification. Such model is more invariant to compositions, and focuses more on the artistic style patterns. We record how many of $I_g$ are classified correctly as the style of $I_{style}$, which reflects how well the generator captures the style features and translates them into $I_g$. The style-classification accuracy for VGG is 95.1\%, indicating a trustworthy performance.

\subsection{Comparison to Baselines:} MUNIT \cite{huang2018multimodal} (unsupervised) and BicycleGAN \cite{zhu2017toward} (supervised) are the two state-of-the-art I2I models that are comparable to our model for their ability to do conditional image translation. SketchyGAN \cite{chen2018sketchygan} is the latest model that is dedicated to the sketch-to-image task. Pix2pix \cite{isola2017image} is the fundamental model for the I2I process. In this section, we show comparisons between our model and the aforementioned models. We also include the results from the classical NST method by \cite{gatys2015neural} as a representative of that class of methods. For SketchyGAN (noted as ``Pix2pix+MRU" since its main contribution is MRU) and Pix2pix, we adopt their model components but use our training schema to enable them the style-conditioned sketch-to-image synthesis.

\begin{figure}
\centering
  \includegraphics[width=1\linewidth,height=0.36\linewidth]{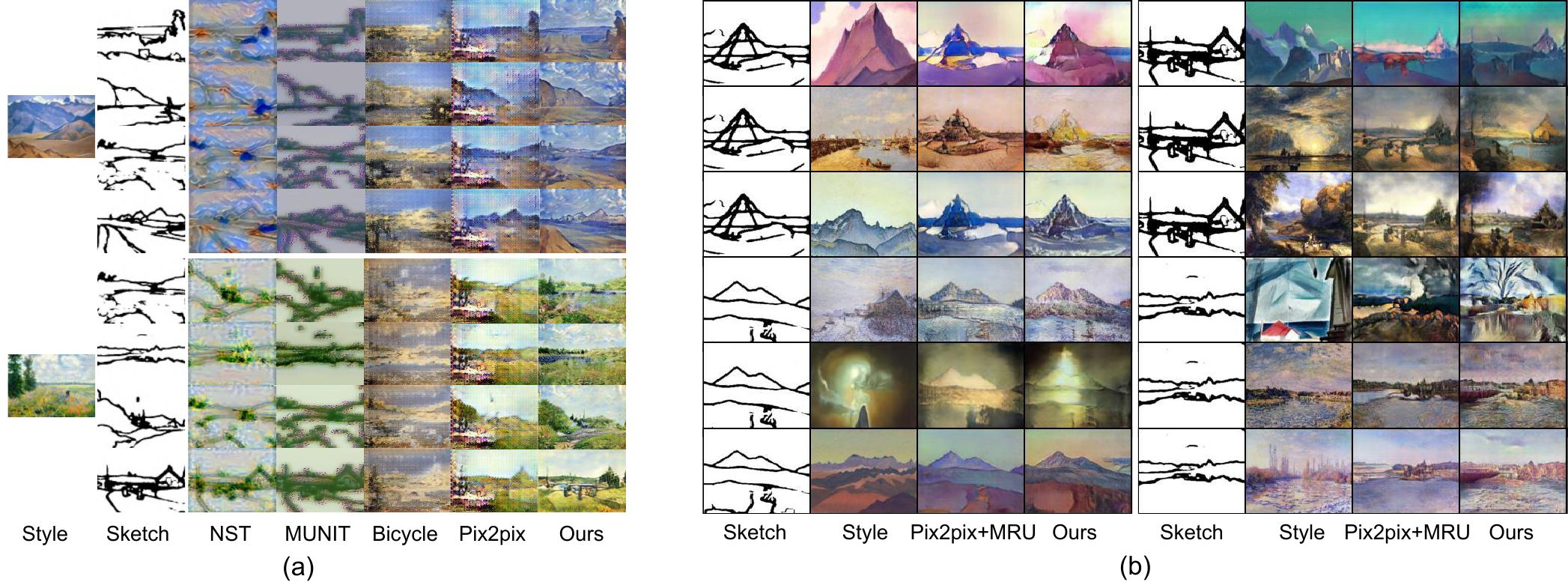}
  \caption{Qualitative comparison to baseline models }
  \label{fig:baselines}
\end{figure}

All the tested models are trained on images with $128\times128$ resolution due to the compared models' capacity restrictions (our model can easily upscale to $512\times512$). We make sure all the compared models have a similar generative capacity by having a similar amount of weights in their respective generators. Except for Pix2pix, all the compared models have more total parameters than ours and require a longer time to train. We tested multiple hyper-parameter settings on baseline models and reported the highest figures.

Qualitative results are shown in Figure~\ref{fig:baselines}. Except ``Pix2pix+MRU", all the other methods can hardly handle the art dataset and do not produce meaningful results. Due to the limited information in sketches, NST and MUNIT can hardly generate meaningful images, BicycleGAN only generates images with blurry edges and fuzzy colors, and Pix2pix consistently gets undesirable artifacts. Notice that, models like MUNIT and BicycleGAN do work well on datasets with simple sketches, where the images share one standard shape (only cats, shoes) and are well registered with white background, and without any artistic features involved (semantically diversified, different color palettes and texture). In contrast, images in art dataset are much more complicated with multiple objects and different compositions, which result in bad performance for these previous models and show the effectiveness of our proposed components.

\begin{table}
   \begin{center}
\caption{Quantitative comparison to baseline models }
         \resizebox{1\columnwidth}{!}{
            \begin{tabular}{c|c|c|c|c|c|c}
            \hline
            & NST & MUNIT & Pix2pix &  BicycleGAN  & Pix2pix+MRU  & Ours     \\
            \hline
            \textbf{FID} $\downarrow$ & 6.85 & 7.43 $\pm$ 0.08 & 7.08 $\pm$ 0.05& 6.39 $\pm$ 0.05     & 5.05 $\pm$ 0.13     & \textbf{4.18} $\pm$ 0.11        \\
            \hline
             \textbf{Classification Score} $\uparrow$ & 0.182 & 0.241 $\pm$ 0.012 & 0.485 $\pm$ 0.012 & 0.321 $\pm$ 0.009 & 0.487 $\pm$ 0.002 & \textbf{0.573} $\pm$ 0.011  \\
            \hline 
            \end{tabular}
        }
    \label{table:baselines}
    \end{center}
\end{table}

The quantitative results, shown in Table~\ref{table:baselines}, concur with the qualitative results. It is worth noticing that, while ``Pix2pix+MRU" generates comparably visual-appealing images, our model outperforms it by a large margin especially in terms of style classification score, which indicates the superiority of our model in translating the right style cues from the style image into the generated images.

\noindent\textbf{Comparison to MRU from SketchyGAN:} Since SketchyGAN is not proposed for the exemplar-based s2i task, the comparison here is not with SketchyGAN as it is, but rather with a modified version with our training schema to suit our problem definition. While the ``Pix2pix+MRU" results look good from a sketch colorization perspective, they are not satisfactory from the artistic style transfer perspective compared with the given style images (eg., texture of flat area, linear vs painterly). As shown in Figure~\ref{fig:baselines}-(b), ``Pix2pix+MRU" tends to produce dull colors on all its generations, with an undesired color shift compared to the style images. In contrast, our results provide more accurate color palette and texture restoration. Apart from the quantitative result, ``Pix2pix+MRU" is outperformed by our model especially in terms of the fine-grained artistic style features, such as color flatness or fuzziness and edge sharpness.

\subsection{Ablation study} We perform ablation studies to evaluate the three proposed components using a customized Pix2pix+MRU model as the baseline. More experiments can be found in the appendix, including a quantitative evaluation of the content and style disentanglement performance of our model, an effectiveness analysis of an image gradient matching loss for better texture generation, and a more detailed comparison between AdaIN and the proposed FMT for style transfer.

In this section, we show both the cumulative comparisons and individual benefits for each modules. When evaluating FMT, we replace the AdaIN layer on lower level image features with FMT, and show better performance of FMT compared to AdaIN. IDN changes the structure of the discriminator, so when validating the effectiveness of IDN, we add an extra baseline model which has a discriminator naively predicting the sketch and style-vector (noted as ``with 2B'') from its two separate convolution branches. 

In Table 2, the proposed components are cumulatively added to the model. FID and classification scores consistently show that each of the proposed components contributes to the generation performance.  

FMT and IDN bring the most significant boost in FID and Classification, respectively. It is worth noticing how the added two branches on the discriminator (with 2B) hurt the performance and how IDN reverses that drawback and further boosts the performance. We hypothesis that naively adding two branches collected conflicting information during training (content and style) and made the models harder to converge. In contrast, IDN neatly eliminates the conflict thanks to its disentangling effect, and takes advantage of both the content and style information for a better generation performance.

\begin{table}[t]
    \begin{center}
        \caption{Ablation study of the proposed components }
        
         \resizebox{1\columnwidth}{!}{
            \begin{tabular}{c|c|c|c|c|c}
                \hline
                & baseline  & with DMI  & with FMT    & with 2B & with IDN     \\
                \hline
                \textbf{FID} $\downarrow$ & 4.77 $\pm$ 0.14   & 4.84  $\pm$ 0.09    & 4.43       $\pm$ 0.15    & 4.73  $\pm$ 0.09  & \textbf{4.18}$\pm$ 0.11  \\
                \hline
                \textbf{Classification Score} $\uparrow$ & 0.485 $\pm$ 0.012    & 0.507  $\pm$ 0.007    & 0.512 $\pm$ 0.009  & 0.479 $\pm$ 0.013  & \textbf{0.573} $\pm$ 0.011 \\
                \hline
            \end{tabular}
        }
    \end{center}
    
    \label{table:ablation}
\end{table}

\begin{figure}[h]
\centering
  \includegraphics[width=0.9\linewidth]{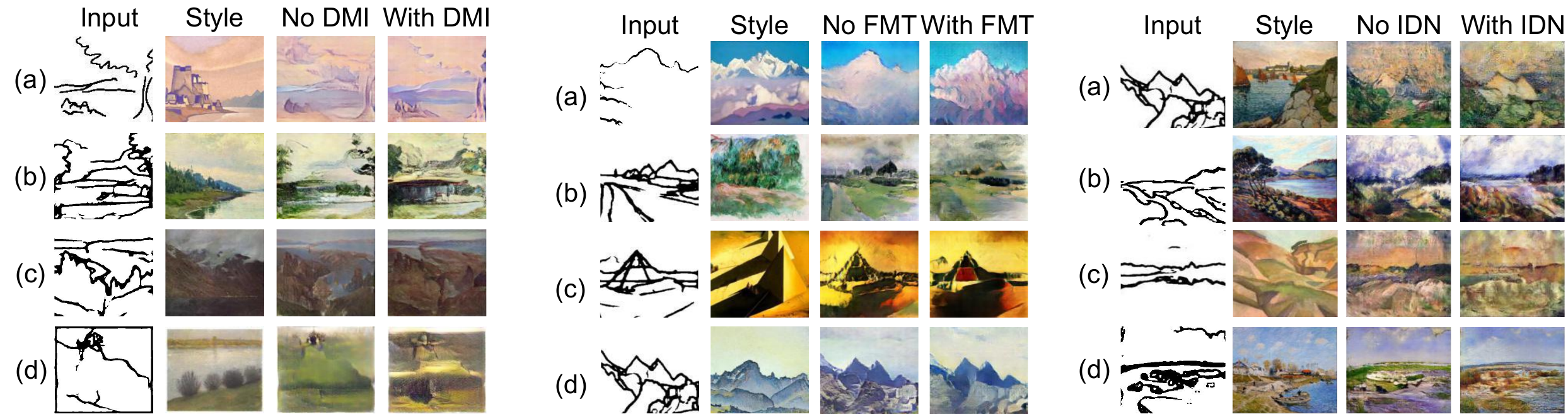}
  \caption{Qualitative comparisons of DMI, FMT and IDN}
  \label{fig:ablation}
\end{figure}

\noindent\textbf{Qualitative Comparisons:} In Figure~\ref{fig:ablation}, the left panel shows how DMI boosts the content faithfulness to the input sketches. In row (a), the model without DMI misses the sketch lines on the top and right portion, while DMI redeems all the lines in $I_{sketch}$. In row (b), $I_g$ without DMI is affected by $I_{style}$ which has a flat and empty sky and misses the lines on the top portion of $I_{sketch}$. In contrast, the model with DMI successfully generates trees, clouds, and mountain views following the lines in $I_{sketch}$. In row (c), $I_g$ without DMI totally messes up the shapes in the mid area in $I_{sketch}$, while, in the one with DMI, all the edges are correctly shaped with clear contrast.

The middle panel in Figure~\ref{fig:ablation} shows how FMT helps translate the style from the input style images. In row (a), FMT helps generate the correct colors and rich textures in the mountain as in $I_{style}$. In row (b), $I_g$ with FMT inherits the smoothness from $I_{style}$ where the colors are fuzzy and the edges are blurry, while removing FMT leads to sharp edges and flat colors. When $I_{style}$ is flat without texture, FMT is also able to resume the correct style. Row (c) and row (d) demonstrate that when $I_{style}$ is clean and flat without fuzziness, $I_g$ without FMT generates undesired textures while FMT ensures the accurate flat style.

The right panel in Figure~\ref{fig:ablation} shows how IDN helps maintain a better color palette and generally reduce the artifacts. In row (a) and (b), there are visible artifacts in the sky of the generated images without IDN, while IDN greatly reduces such effects. In row (b) and (c), the images without IDN shows undesired green and purple-brown colors that are not in the given style images, and IDN has better color consistency. In row (c) and (d), there are clear color-shifts in the generated images, which then been corrected in the model with IDN. Please refer to the Appendix for more qualitative comparisons.

\subsection{Human Evaluation}
Compared with the ``Pix2pix+MRU" baseline model, we conduct human survey to validate our model's effectiveness. The survey is taken by 100 undergraduate students to ensure the quality. In each question, the user is presented with a sketch, a style image and the generated images from baseline (Pix2pix+MRU) and our model (option letters are randomly assigned to reduce bias) and asked: ``Which image, a or b, better captures the style in colorizing the sketch?". We collected 1000 results and 63.3\% selects our model as the better performer. 13.3\% selects ``hard to tell". Only 23.4\% prefer the baseline model.

\subsection{Qualitative Results on Multi Domains}
\begin{figure}[h]
\centering
  \includegraphics[width=0.9\linewidth,height=0.4\linewidth]{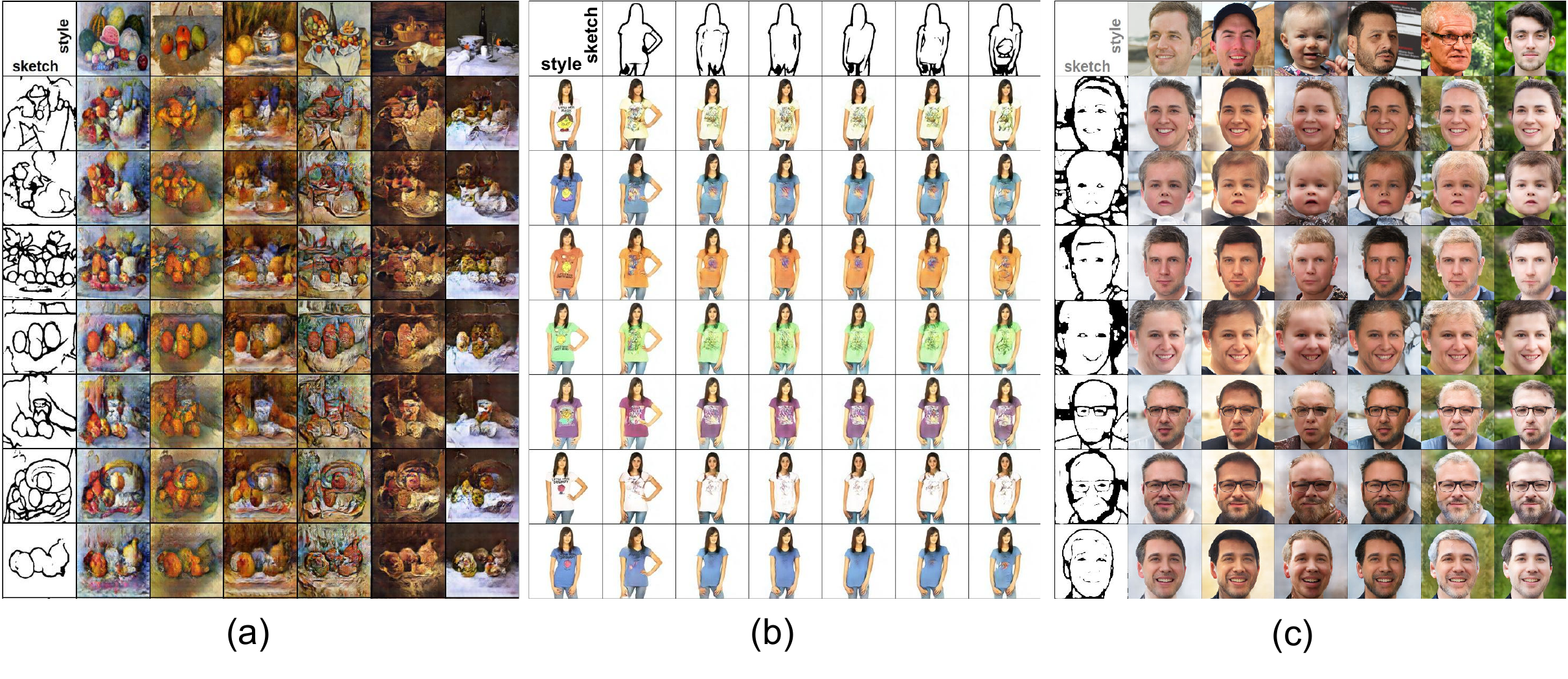}
  \caption{Synthesize on still-life painting, apparel, and portrait at $512^2$ resolution.}
  \label{fig:multi-genres}
\end{figure}

While focused on art, our model is generic to other image domains with superior visual quality than previous models. On more commonly used datasets CelebA and Fashion Apparel, our model also out-performs the baselines and shows the new state-of-the-art performance. Figure~\ref{fig:multi-genres} shows the results on multiple image domains. In figure (c), the glasses in the sketches are successfully rendered in the generated images even when there is no glasses in the style image. Similarly, the moustache is correctly removed when there is moustache in the style images but is not indicated in the sketches. It's worth noticing that the gender of generated images follows the sketch instead of style. These results show clear evidence that the model learns semantics of input sketch from the training corpus.

\section{Discussions}
The model yields consistent performance when we try different input settings for the synthesis, including using multiple reference style images rather than one. During the experiments, we also discovered some interesting behaviors and limitations that worth further research. For instance, learning from the corpus can be a good thing for providing extra style cues apart from the style image, however, it may also cause conflicts against it, such as inaccurate coloring and excess shapes. It is worth study on how to balance the representation the model learns from the whole corpus and from the referential style image, and how to take advantage of the knowledge from the corpus for better generation. We sincerely guide the readers to the Appendix for more information.

\section{Conclusion}
We approached the task of generating artistic images from sketch while conditioned on style images with a novel sketch-to-art model. Unlike photo-realistic datasets, we highlighted and identified the unique properties of artistic images and pointed out the different challenges they possess. Respectively, we proposed methods that can effectively addressed these challenges. Our model synthesizes images with the awareness of more comprehensive artistic style attributes, which goes beyond color palettes, and for the first time, identifies the varied texture, contours, and plain area styles. Overall, our work pushes the boundary of the deep neural networks in capturing and translating various artistic styles, and makes a solid contribution to the sketch-to-image literature.

\noindent\textbf{Acknowledgement} At \url{https://create.playform.io/sketch-to-image}, demo of the model in this paper is available. The research was done while Bingchen Liu, Kunpeng Song and Ahmed Elgammal were at Artrendex Inc. 

\clearpage

\bibliographystyle{splncs}
\bibliography{egbib}

\begin{thebibliography}{10}

\bibitem{elgammal2017can}
Elgammal, A., Liu, B., Elhoseiny, M., Mazzone, M.:
\newblock Can: Creative adversarial networks, generating" art" by learning
  about styles and deviating from style norms.
\newblock arXiv preprint arXiv:1706.07068 (2017)

\bibitem{elgammal2018shape}
Elgammal, A., Mazzone, M., Liu, B., Kim, D., Elhoseiny, M.:
\newblock The shape of art history in the eyes of the machine.
\newblock arXiv preprint arXiv:1801.07729 (2018)

\bibitem{kim2018finding}
Kim, D., Liu, B., Elgammal, A., Mazzone, M.:
\newblock Finding principal semantics of style in art.
\newblock In: 2018 IEEE 12th International Conference on Semantic Computing
  (ICSC), (IEEE)  156--163

\bibitem{chen2018sketchygan}
Chen, W., Hays, J.:
\newblock Sketchygan: Towards diverse and realistic sketch to image synthesis.
\newblock In: CVPR. (2018)  9416--9425

\bibitem{zhang2017style}
Zhang, L., Ji, Y., Lin, X., Liu, C.:
\newblock Style transfer for anime sketches with enhanced residual u-net and
  auxiliary classifier gan.
\newblock In: ACPR. (2017)  506--511

\bibitem{schapiro1994theory}
Schapiro, M., Schapiro, M., Schapiro, M., Schapiro, M.:
\newblock Theory and philosophy of art: Style, artist, and society. Volume~4.
\newblock George Braziller New York (1994)

\bibitem{sangkloy2017scribbler}
Sangkloy, P., Lu, J., Fang, C., Yu, F., Hays, J.:
\newblock Scribbler: Controlling deep image synthesis with sketch and color.
\newblock In: CVPR. (2017)  5400--5409

\bibitem{gatys2015neural}
Gatys, L.A., Ecker, A.S., Bethge, M.:
\newblock Image style transfer using convolutional neural networks.
\newblock In: CVPR. (2016)  2414--2423

\bibitem{goodfellow2014generative}
Goodfellow, I., Pouget-Abadie, J., Mirza, M., Xu, B., Warde-Farley, D., Ozair,
  S., Courville, A., Bengio, Y.:
\newblock Generative adversarial nets.
\newblock In: NIPS. (2014)  2672--2680

\bibitem{ulyanov2016instance}
Ulyanov, D., Vedaldi, A., Lempitsky, V.:
\newblock Instance normalization: The missing ingredient for fast stylization.
\newblock arXiv preprint arXiv:1607.08022 (2016)

\bibitem{isola2017image}
Isola, P., Zhu, J.Y., Zhou, T., Efros, A.A.:
\newblock Image-to-image translation with conditional adversarial networks.
\newblock In: CVPR. (2017)  1125--1134

\bibitem{zhu2017unpaired}
Zhu, J.Y., Park, T., Isola, P., Efros, A.A.:
\newblock Unpaired image-to-image translation using cycle-consistent
  adversarial networks.
\newblock In: ICCV. (2017)  2223--2232

\bibitem{liu2017unsupervised}
Liu, M.Y., Breuel, T., Kautz, J.:
\newblock Unsupervised image-to-image translation networks.
\newblock In: NIPS. (2017)  700--708

\bibitem{choi2018stargan}
Choi, Y., Choi, M., Kim, M., Ha, J.W., Kim, S., Choo, J.:
\newblock {StarGAN}: Unified generative adversarial networks for multi-domain
  image-to-image translation.
\newblock In: CVPR. (2018)  8789--8797

\bibitem{yi2017dualgan}
Yi, Z., Zhang, H., Tan, P., Gong, M.:
\newblock Dualgan: Unsupervised dual learning for image-to-image translation.
\newblock In: ICCV. (2017)  2849--2857

\bibitem{huang2018multimodal}
Huang, X., Liu, M.Y., Belongie, S., Kautz, J.:
\newblock Multimodal unsupervised image-to-image translation.
\newblock In: ECCV. (2018)  172--189

\bibitem{zhu2017toward}
Zhu, J.Y., Zhang, R., Pathak, D., Darrell, T., Efros, A.A., Wang, O.,
  Shechtman, E.:
\newblock Toward multimodal image-to-image translation.
\newblock In: NIPS. (2017)  465--476

\bibitem{almahairi2018augmented}
Almahairi, A., Rajeswar, S., Sordoni, A., Bachman, P., Courville, A.:
\newblock Augmented {cycleGAN}: Learning many-to-many mappings from unpaired
  data.
\newblock In: ICML. (2018)  195--204

\bibitem{lee2018diverse}
Lee, H.Y., Tseng, H.Y., Huang, J.B., Singh, M., Yang, M.H.:
\newblock Diverse image-to-image translation via disentangled representations.
\newblock In: ECCV. (2018)  35--51

\bibitem{liu2020time}
Liu, B., Song, K., Zhu, Y., de~Melo, G., Elgammal, A.:
\newblock Time: Text and image mutual-translation adversarial networks.
\newblock arXiv preprint arXiv:2005.13192 (2020)

\bibitem{zhu2020s3vae}
Zhu, Y., Min, M.R., Kadav, A., Graf, H.P.:
\newblock S3vae: Self-supervised sequential vae for representation
  disentanglement and data generation.
\newblock In: Proceedings of the IEEE/CVF Conference on Computer Vision and
  Pattern Recognition. (2020)  6538--6547

\bibitem{johnson2016perceptual}
Johnson, J., Alahi, A., Fei-Fei, L.:
\newblock Perceptual losses for real-time style transfer and super-resolution.
\newblock In: ECCV. (2016)  694--711

\bibitem{zhang2017multistyle}
Zhang, H., Dana, K.:
\newblock Multi-style generative network for real-time transfer.
\newblock In: ECCV. (2018)

\bibitem{huang2017arbitrary}
Huang, X., Belongie, S.:
\newblock Arbitrary style transfer in real-time with adaptive instance
  normalization.
\newblock In: ICCV. (2017)  1501--1510

\bibitem{jing2017neural}
Jing, Y., Yang, Y., Feng, Z., Ye, J., Yu, Y., Song, M.:
\newblock Neural style transfer: A review.
\newblock arXiv preprint arXiv:1705.04058 (2017)

\bibitem{liu2017auto}
Liu, Y., Qin, Z., Luo, Z., Wang, H.:
\newblock Auto-painter: Cartoon image generation from sketch by using
  conditional generative adversarial networks.
\newblock arXiv preprint arXiv:1705.01908 (2017)

\bibitem{zhang2018two}
Zhang, L., Li, C., Wong, T.T., Ji, Y., Liu, C.:
\newblock Two-stage sketch colorization.
\newblock In: SIGGRAPH Asia 2018 Technical Papers. (2018)  261

\bibitem{park2019semantic}
Park, T., Liu, M.Y., Wang, T.C., Zhu, J.Y.:
\newblock Semantic image synthesis with spatially-adaptive normalization.
\newblock In: CVPR. (2019)  2337--2346

\bibitem{ronneberger2015u}
Ronneberger, O., Fischer, P., Brox, T.:
\newblock {U-Net}: Convolutional networks for biomedical image segmentation.
\newblock In: International Conference on Medical image computing and
  computer-assisted intervention. (2015)  234--241

\bibitem{kazemi2019style}
Kazemi, H., Iranmanesh, S.M., Nasrabadi, N.:
\newblock Style and content disentanglement in generative adversarial networks.
\newblock In: WACV. (2019)  848--856

\bibitem{karras2019style}
Karras, T., Laine, S., Aila, T.:
\newblock A style-based generator architecture for generative adversarial
  networks.
\newblock In: CVPR. (2019)  4401--4410

\bibitem{wikiart}
:
\newblock Wikiart.
\newblock (https://www.wikiart.org/)

\bibitem{heusel2017gans}
Heusel, M., Ramsauer, H., Unterthiner, T., Nessler, B., Hochreiter, S.:
\newblock {GANs} trained by a two time-scale update rule converge to a local
  nash equilibrium.
\newblock In: NIPS. (2017)  6626--6637

\bibitem{russakovsky2015imagenet}
Russakovsky, O., Deng, J., Su, H., Krause, J., Satheesh, S., Ma, S., Huang, Z.,
  Karpathy, A., Khosla, A., Bernstein, M.,  et~al.:
\newblock Imagenet large scale visual recognition challenge.
\newblock IJCV \textbf{115} (2015)  211--252

\bibitem{he2016deep}
He, K., Zhang, X., Ren, S., Sun, J.:
\newblock Deep residual learning for image recognition.
\newblock In: CVPR. (2016)  770--778

\bibitem{hu2018squeeze}
Hu, J., Shen, L., Sun, G.:
\newblock Squeeze-and-excitation networks.
\newblock In: CVPR. (2018)  7132--7141

\bibitem{jing2019neural}
Jing, Y., Yang, Y., Feng, Z., Ye, J., Yu, Y., Song, M.:
\newblock Neural style transfer: A review.
\newblock IEEE transactions on visualization and computer graphics (2019)

\end{thebibliography}

\clearpage

\begin{center}
	\textbf{\Large Appendix \\
	 Sketch-to-Art: Synthesizing Stylized Art Images From Sketches}
\end{center}

\appendix

\section{More Qualitative Results}

\begin{figure}[h]
    \centering
    \includegraphics[width=1\linewidth,height=0.5\linewidth]{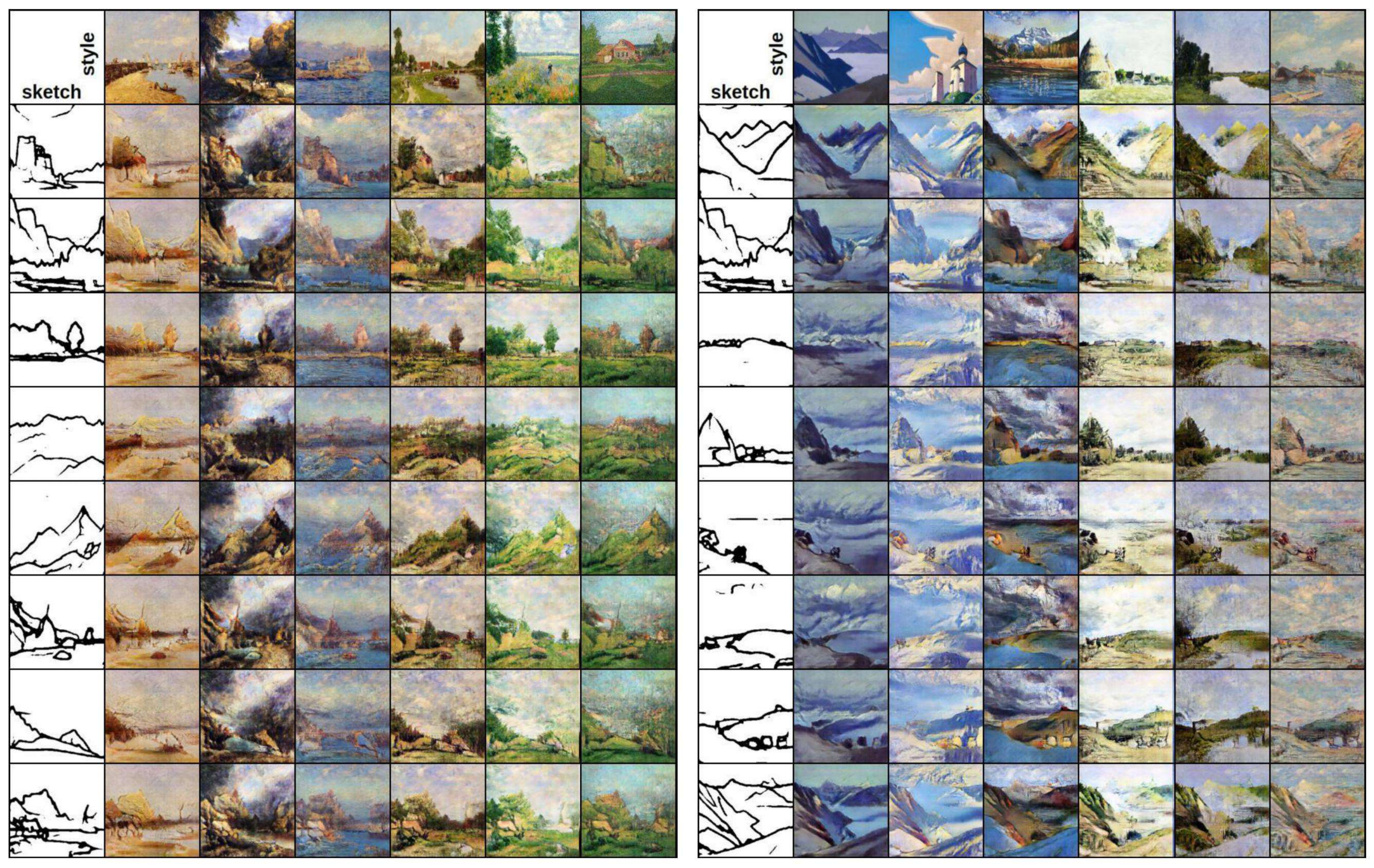}
    \caption{Qualitative results from our model trained on genre \textbf{landscape}}
    \label{fig:arb_1}
\end{figure}

\begin{figure}[h]
    \centering
    \includegraphics[width=\linewidth,height=0.5\linewidth]{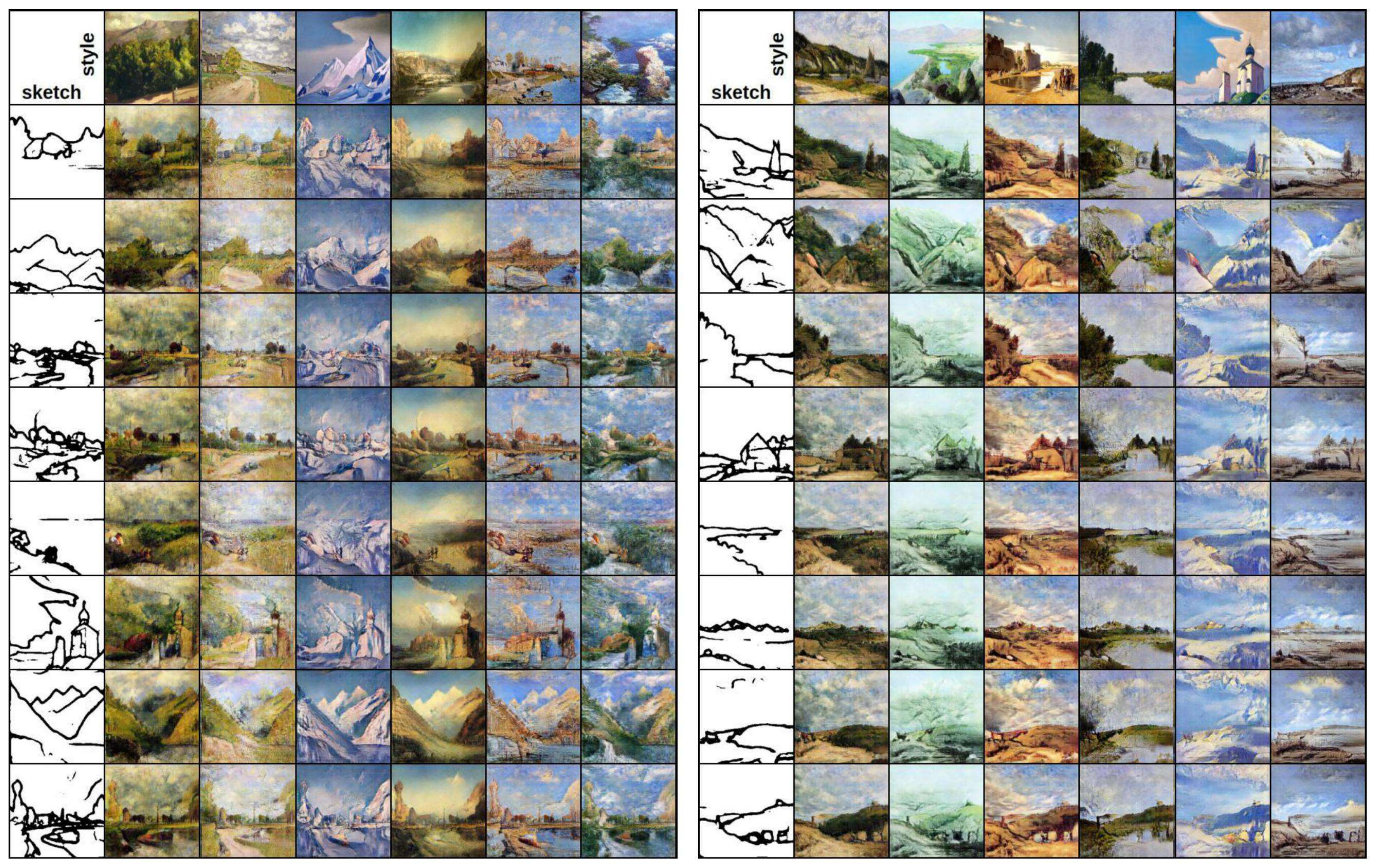}
    \caption{Qualitative results from our model trained on genre \textbf{landscape}}
    \label{fig:arb_2}
\end{figure}

\begin{figure}[h]
    \centering
    \includegraphics[width=1\linewidth]{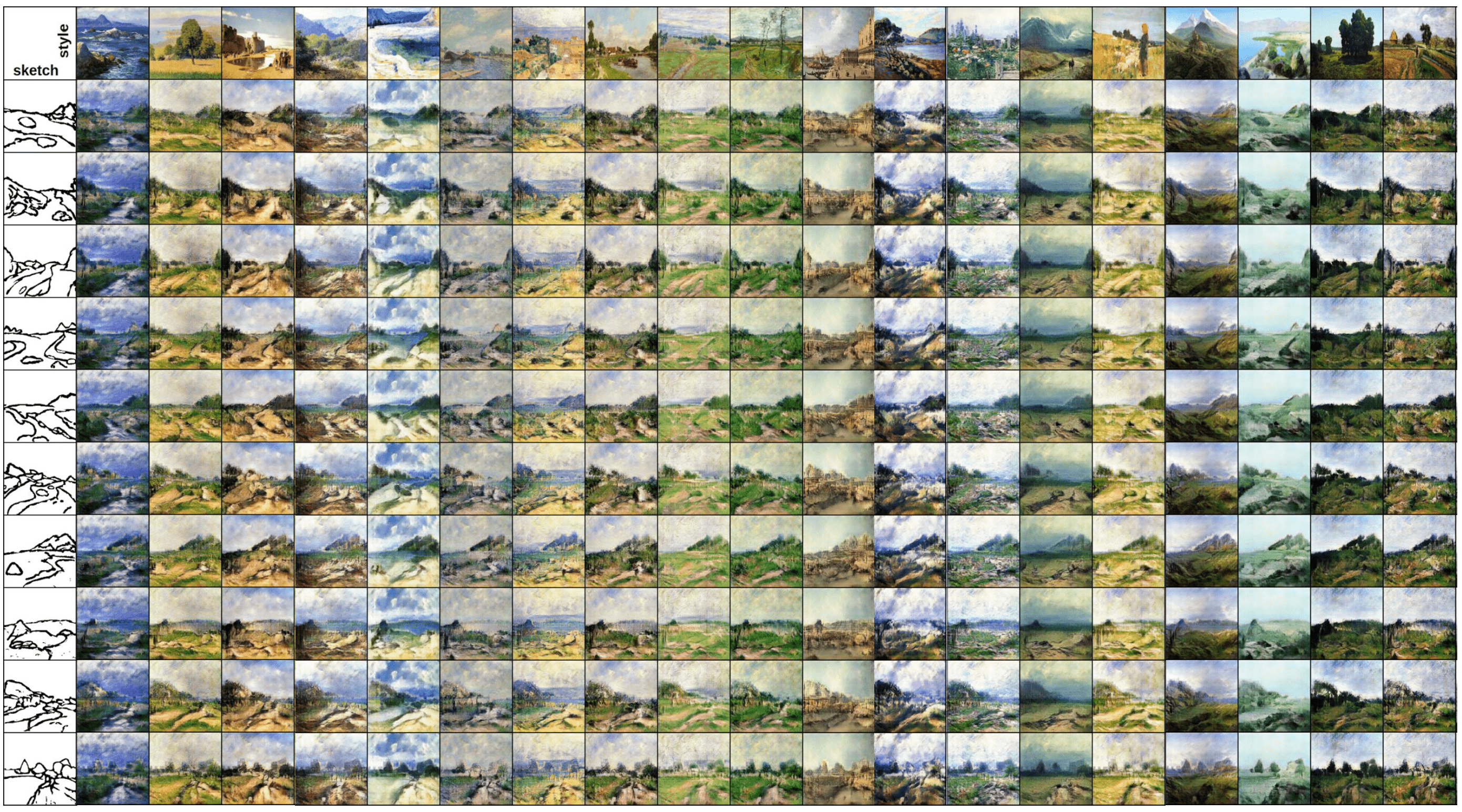}
    \caption{\textbf{Synthesizing with hand-draw sketches}}
    \label{fig:hand_draw_1}
\end{figure}
\begin{figure}[h]
    \centering
    \includegraphics[width=1\linewidth]{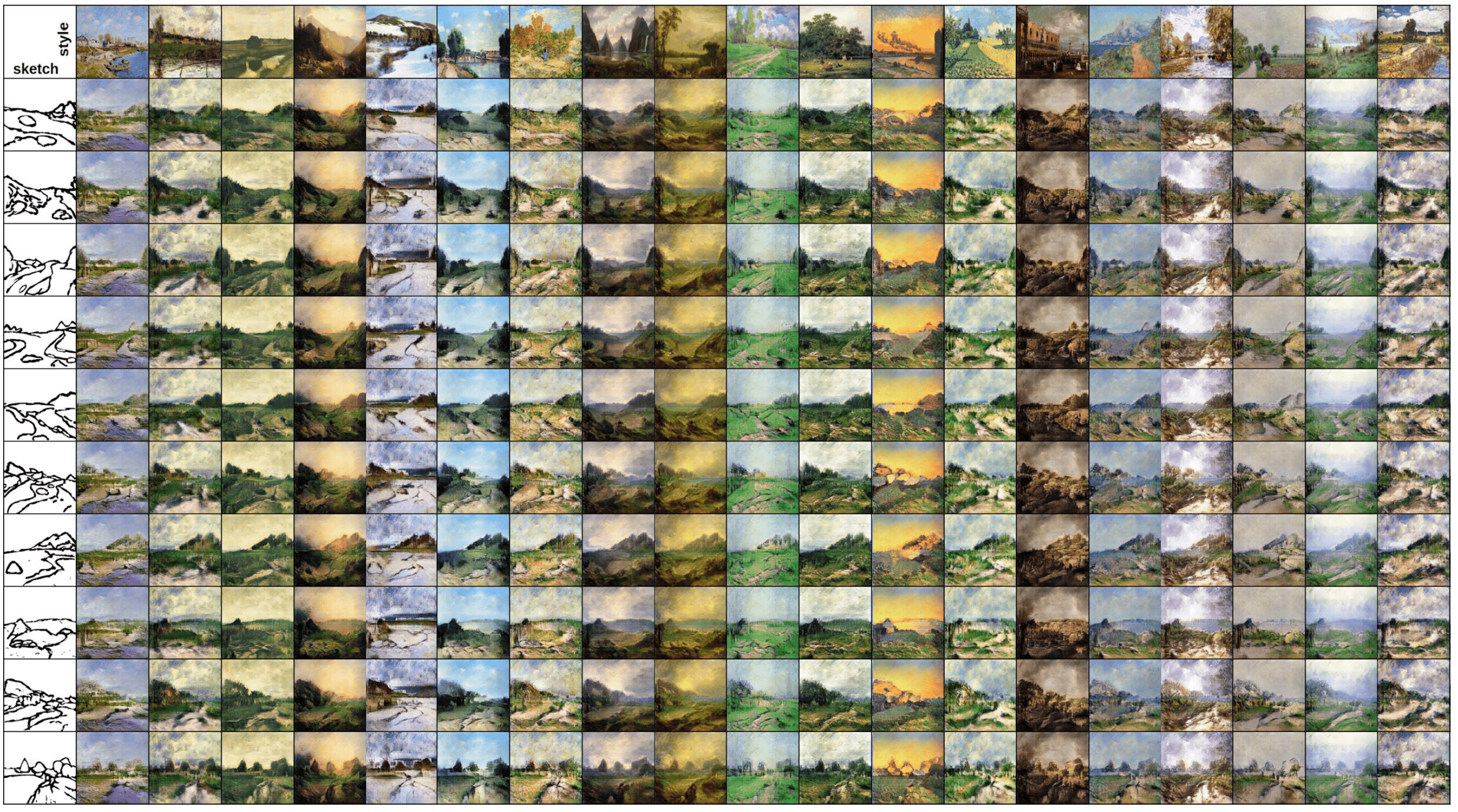}
    \caption{\textbf{Synthesizing with hand-draw sketches}}
    \label{fig:hand_draw_2}
\end{figure}

\begin{figure}[h]
    \centering
    \includegraphics[width=\linewidth]{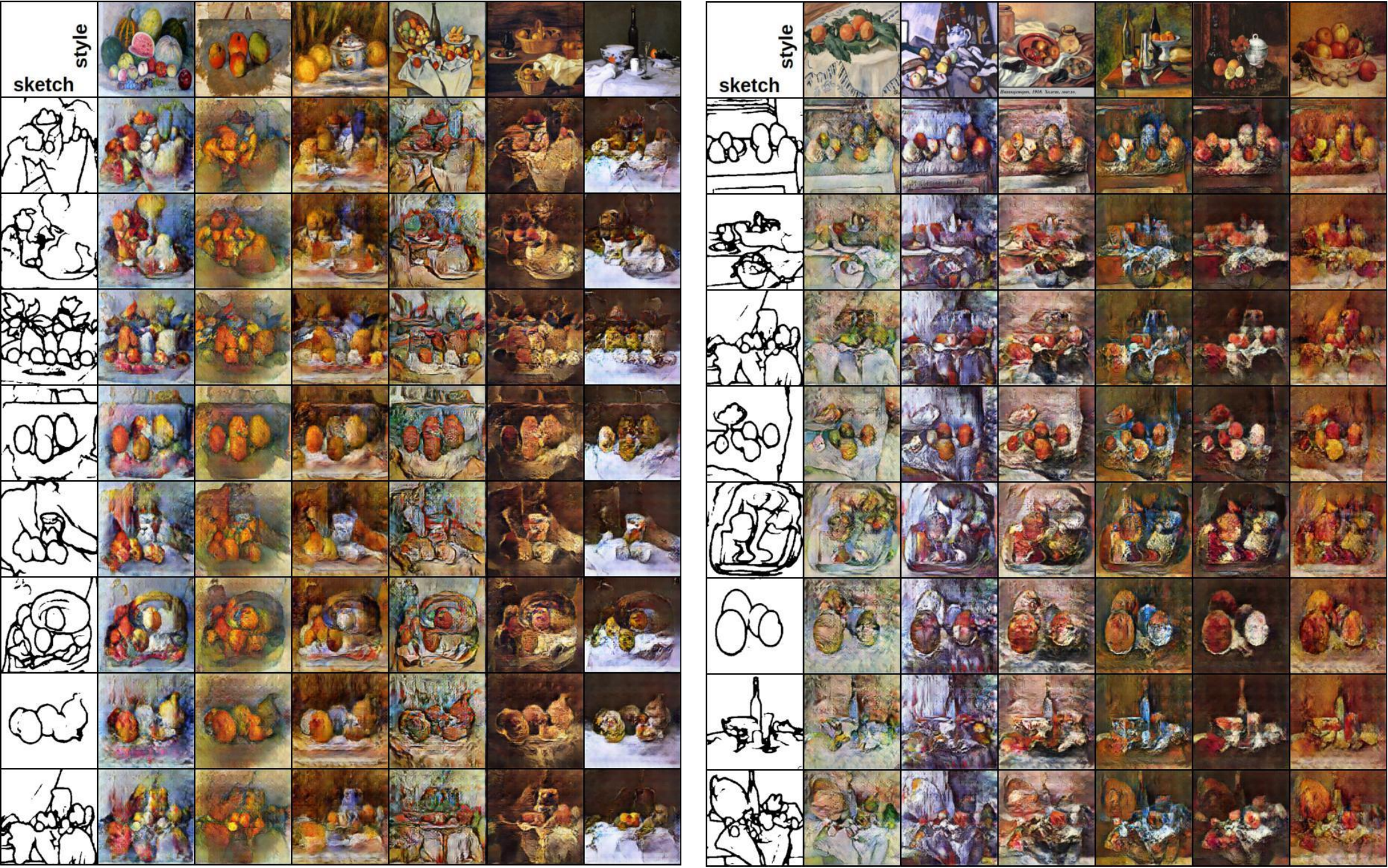}
    \caption{Qualitative results from our model trained on genre \textbf{still-life}}
    \label{fig:arb_3}
\end{figure}

\section{Model Components}

\subsection{Attention-based Residual Block}
\textbf{Description: }
We customize the convolution structure in $G$ and $D$ based on the residual blocks proposed by \cite{he2016deep}. Specifically, we apply a channel-wise attention following \cite{hu2018squeeze} on the second convolution layer in each block. To our knowledge, we are the first to adopt such attention-based layer within residual blocks. This tweak brings significant image quality boost in our task from the convolution layers used in Pix2Pix and BicycleGAN \cite{isola2017image,zhu2017toward} while maintains minimum extra computing cost. In sketch-to-image task, traditional convolution layers or residual convolutions suffer from fuzzy artifices and can hardly generate diverse colors. The proposed attention-based residual block largely improved such scenario for the baseline models. All the experimental results we present in this paper are based on this tweak. Further experiments are required to validate its effectiveness in general tasks, however, it is an orthogonal component that is beyond the discussion scope of this paper.

\begin{figure}[h]
    \centering
    \includegraphics[width=0.5\linewidth, height=5cm]{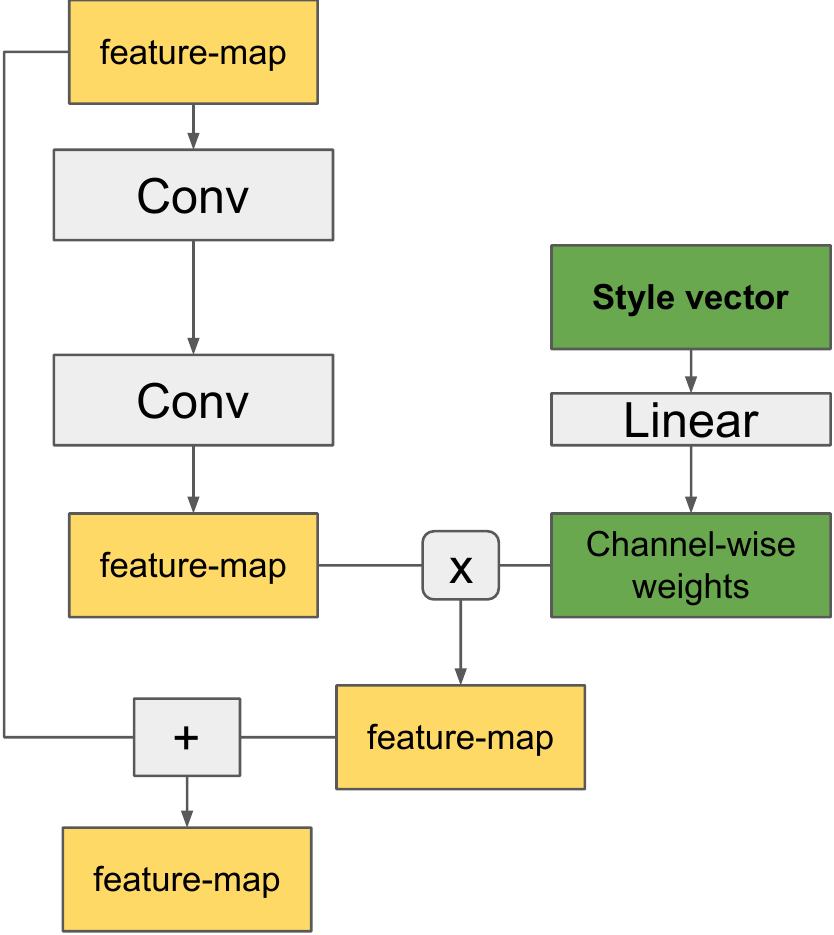}
    \caption{Attention-based residual block}
    \label{fig:attn_res_blk}
\end{figure}

The tweak of the attention-based residual block is illustrated in Figure~\ref{fig:attn_res_blk}. It consists of two convolution layers and one fully-connected layer (linear layer). It takes two input, one is the feature-map $f$ from the previous layer, and the other one is an style vector $V_{style}$ from our feature extractor $E$. During training, the two convolution layers will compute a new feature-map $f'$, while the linear layer will take $V_{style}$ as input and output a channel-wise weight $w_{style}$. Unlike traditional residual block which directly add $f'$ back to $f$, $w_{style}$ will provide the weights to scale the values in each channel of $f'$ before the add-back operation. 

\noindent\textbf{Intuition: }We assume that different channels in the feature-map carry the information that corresponding to different artistic styles. If we have a weight vector that can control the attendance of each channel, i.e. mute some channels' value to zero and amplify some other channels' value, it will make the residual block easier to learn a diversified features. Also, the extra style vector provides more information during the generation process. The generative power is therefore largely increased.

During training, the extra linear layer in the residual block introduces almost none extra computing time. However, it makes the model much faster to converge (the model us able to generate meaningful images usually after just one epoch of training), and also results in much better final performance.

\subsection{Image Gradient Matching Loss} 
A critical element of style is whether contours are linear (sharp) or painterly and whether planer areas are flat or textured. To capture these differences, we propose a Gradient Matching loss that integrates statistics about image gradient. Specifically, we match the gradient statistics of the generated image and their feature maps to the ones of the style image as a new training objective.

We use a patched version of gradient matching on image level and also on the conv-layer activation level. For a given input $I$ and a target $T$ where ($I$, $T$)$\in{\rm I\!R}^{C \times H \times W}$, we first divide them spatially into $8 \times 8$ patches and  compute the gradient loss within each patch, then we will average among all patches to get the final loss. To ignore the content differences between the generated image and target image, we match the mean and variance of the gradients in each patch instead of directly matching by pixels as follows:
\begin{align}\label{eq:gm}
\begin{aligned}
\mathcal{L}_{gradient}= \frac{1}{n} \sum_{p=0}^{n} & (||\mu(\nabla{I_p}) -\mu(\nabla{T_p})||^2 \\
& + ||\sigma(\nabla{I_p}) - \sigma(\nabla{T_p})||^2).
\end{aligned}
\end{align}
The intuition of using gradient matching to capture style features is similar to gram-matrix matching \cite{gatys2015neural,zhang2017multistyle,jing2019neural}. However, gram-matrix carries more of the color palette information and even some shape information, and usually leads to similar yet artificial textures at unwanted locations. Gradient matching preserves the diversity and is more representative on various style textures, such as the brush stroke styles and contour sharpness and fuzziness. 

\begin{table}[t]
    \caption{FID for the gradient-matching loss. }\smallskip
    
    \centering
    \resizebox{1\columnwidth}{!}{
        \smallskip
            \begin{tabular}{c c c c c}
            \midrule 
                 & baseline (Pix2pix+MRU)  & ours w/o G-loss  & ours with G-loss  & ours with Gram-matrix     \\
            \midrule
            mean & 4.77     & 4.43               & \textbf{4.28}         & 4.51         \\
            std  & 0.14     & 0.15               & 0.04         & 0.09   \\
            \hline
            \end{tabular}
    }
    \label{table:g-loss-ablation}
\end{table}

Table~\ref{table:g-loss-ablation} shows the effectiveness of the gradient matching loss, we also compared it with the loss that matches the gram matrix of the images in the same manner. From our experiments on data from other domains, including photo-realistic human faces and fashion apparels, this image gradient matching loss does not improve the result. It is due to the fact that images from those domains do not have a comparable gradient variance as arts with various styles. Instead, they share the same gradient statistics over the whole dataset, thus gradient matching has no effect.

\subsection{The Style and Content Disentanglement}

Our Dual-Mask Injection (DMI) and Instance De-Normalization (IDN) modules are designed to strengthen the content faithfulness of the generated images to the given sketch, which further lead to a better content and style separation capability. To quantitatively show the effectiveness of the proposed modules, we took the edge maps of a set of images as input sketches, paired them with random style images and then extracted the new edge maps from the generated images. In the end, regardless of the style differences, the edge maps from the original image and the generated image should match as much as possible. And a better matched edge map indicates a better content faithfulness.

\begin{table}[h]
\vspace{-5mm}
    \begin{center}
        \caption{Content faithfulness comparison for DMI and IDN}
         \resizebox{1\columnwidth}{!}{
            \begin{tabular}{c c c c c}
                \hline

        & baseline  & with DMI only & with IDN only & with DMI and IDN     \\
                \hline
    \textbf{L1} ($\times$1000) $\downarrow$ &  4.1   & 1.8 &  2.7  & 1.3   \\
                \hline
    \textbf{PDAR} $\downarrow$ &  0.23   & 0.13 &  0.21  & 0.11   \\
                \hline
            \end{tabular}
        }
    \end{center}
    
    \label{table:content_compare}
 \vspace{-0.3cm}
\end{table}

As shown in Table 2, we compute the L1-distance and the Pixel Disagreement Ratio to evaluate the edge map consistency between the input ``sketch" and the output stylized image. DMI makes the biggest performance boost and IDN also contributes to a better style and content separation.

\subsection{Comparison between FMT and AdaIN}
\begin{figure}[h]
    \centering
    \includegraphics[width=0.7\linewidth, height=5cm]{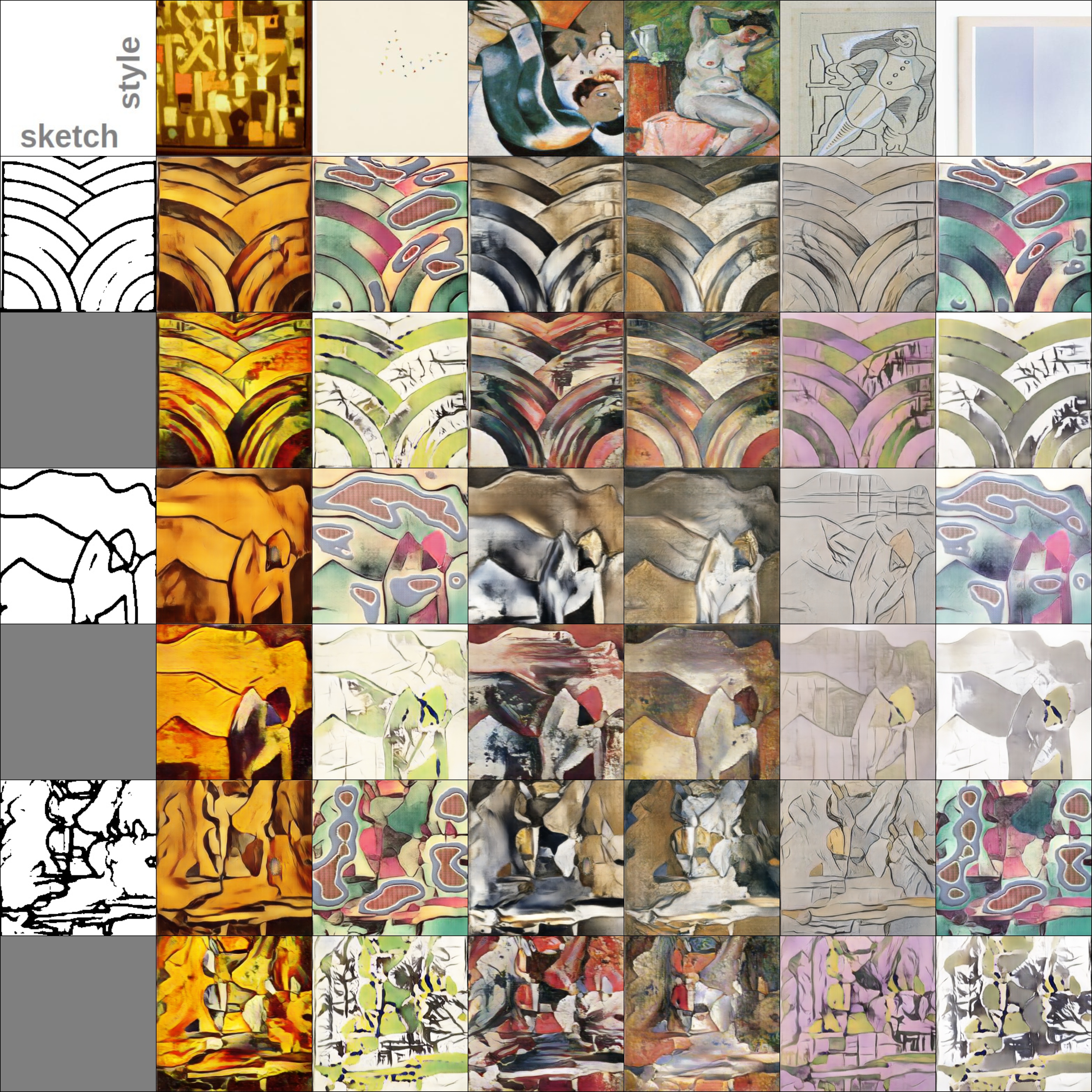}
    \vspace{-2mm}
    \caption{Defects caused by AdaIN. For each sketch, the first row shows the generated images with AdaIN applied on early feature-maps, the second row are samples from model which replace AdaIN with FMT.}
    \label{fig:adain_defect}
    \vspace{-5mm}
\end{figure}

Adaptive Instance Normalization approaches the style transfer task by shifting the source image's feature statistics to match those of the target image. However, while AdaIN is effective for the stylization task of one or a few style images, it has one undesired side-effect when applied on models that are trained on large corpus of images of various styles. As shown in Figure~\ref{fig:adain_defect}, the model trained with AdaIN exhibits artifact patterns that do not belong to any style images. Such defect caused by AdaIN is also discovered in \cite{karras2020analyzing}, where the authors hypothesis the reason as the Generator deliberately seeking to ``fake" a region of high signals in the feature-maps in order to compliment the normalized matching requirements of AdaIN.

\begin{table}[h]
\vspace{-5mm}
    \begin{center}
        \caption{Style transfer comparison between AdaIN and FMT}
         \resizebox{1\columnwidth}{!}{
            \begin{tabular}{c c c c c}
                \hline

        & AdaIn at 16,32,64  & FMT as 16, AdaIn at 32,64  & FMT at 16, 32, AdaIn at 64 & FMT at 16,32,54     \\
                \hline
    \textbf{Gram matrix L2} ($\times$1000) $\downarrow$ & 2.35 $\pm$ 1.4   & 2.14  $\pm$ 0.9    &  1.83 $\pm$  1.2    & 1.47  $\pm$ 0.9   \\

                \hline
            \end{tabular}
        }
    \end{center}
    
    \label{table:adain_fmt_compare}
 \vspace{-0.3cm}
\end{table}

Unlike AdaIN, our proposed Feature-Map-Transformation (FMT) does not apply any force to manipulating the features generated by the Generator. Instead, it provides the ``style" information by concatenating a supplementary feature map from the style images, making the generating process unlikely to have the defect. According to our experiments, no such defect is discovered in our model trained with FMT. Moreover, the style transfer performance is also boosted according to the style classification experiment in the main paper when replacing the AdaIN with FMT. Here in Table 3 we provide a more dedicated experiment by measuring the L2 distance between the gram matrix of the generated images and the style images, over 10000 generated samples. We first use AdaIN on all layers of feature-map at resolution of $16^2, 32^2, 64^2$, then we gradually replace AdaIN by FMT. At each replacement, we can see a more consistent gram matrix between the generated images and the style images, indicating a better aligned style statistics.

\subsection{More Ablation Samples for IDN}
Unlike DMI and FMT, IDN improves the overall generative quality by enabling the Discriminator with the ability to capture the essential content and style features. It is hard to intuitively and selectively point out one direction IDN mainly focuses on. So apart from the FID score, we put more generated samples to qualitatively demonstrate the benefits of it. According to our observation, we conclude two major aspects of IDN's contribution: 1. less artifacts(notice the sky part) and more re-fined image synthesis, and 2. more accurate color consistency to the referential style image and no color-shift effects (notice the 3rd, 6th, 8th style).

\begin{figure}[h]
    \centering
    \includegraphics[width=1\linewidth, height=6cm]{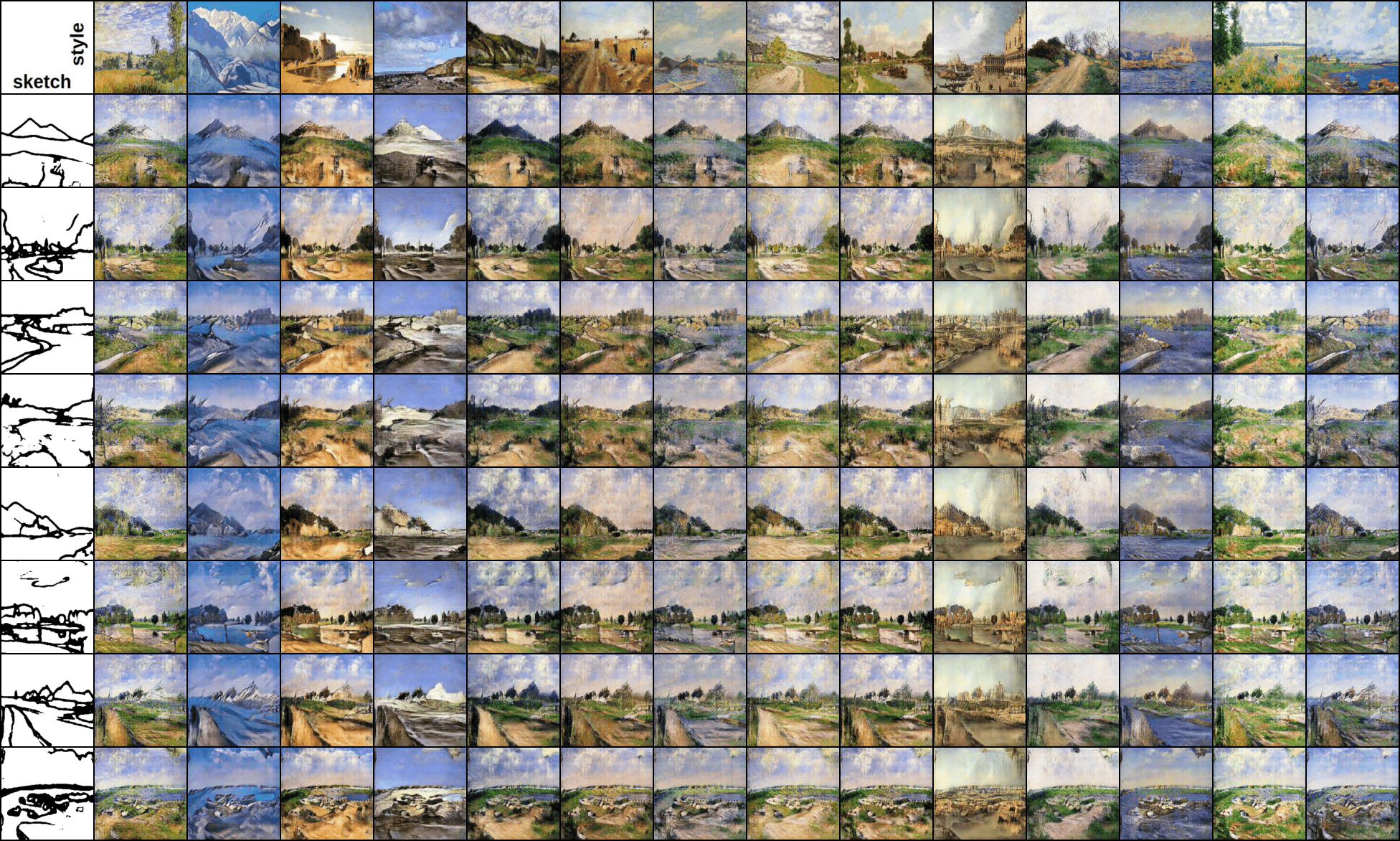}
    \vspace{-5mm}
    \caption{Generated images without IDN module}
    \label{fig:no-idn}
    \vspace{5mm}
    \centering
    \includegraphics[width=1\linewidth, height=6cm]{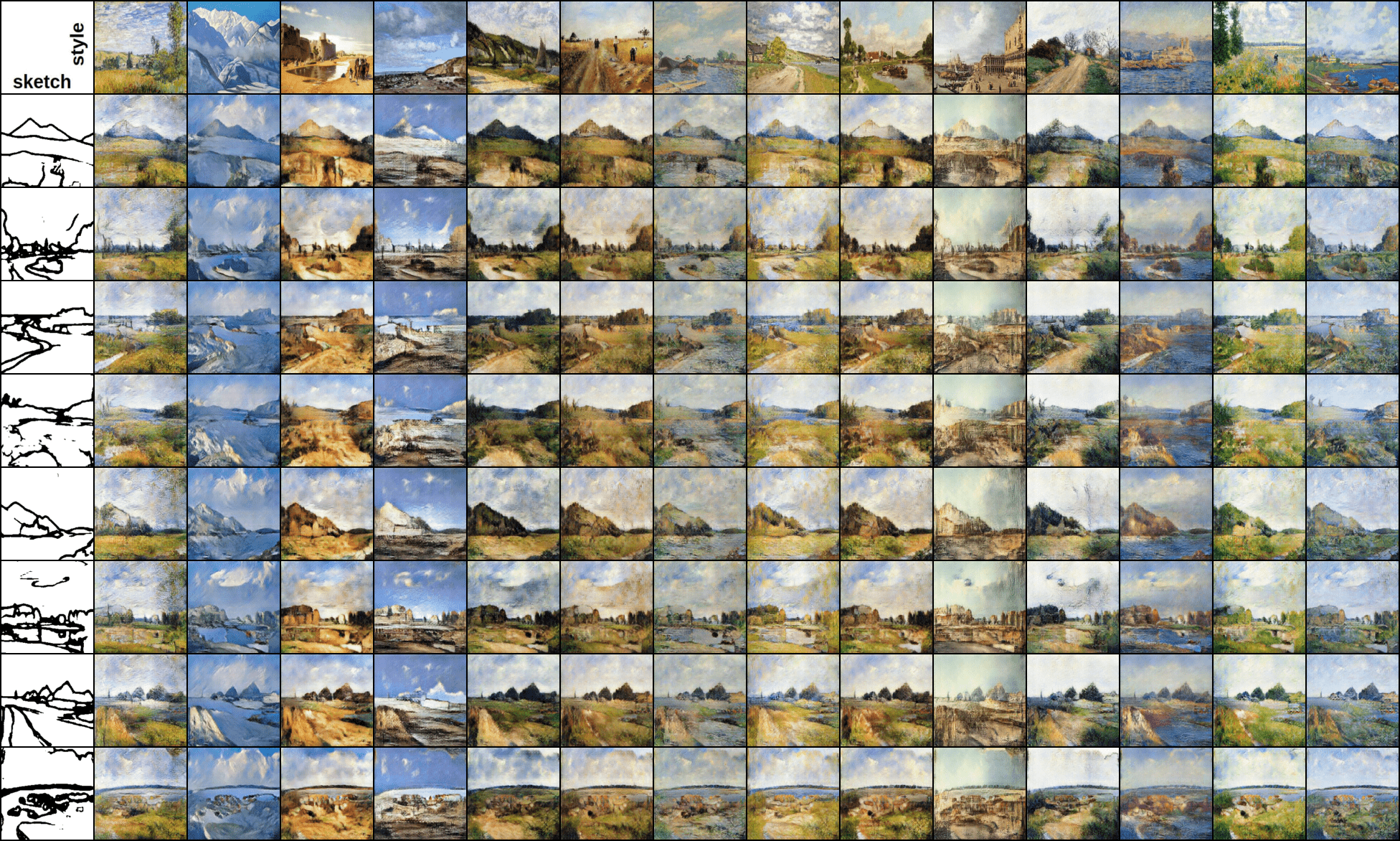}
    \vspace{-5mm}
    \caption{Generated images with IDN module}
    \label{fig:with-idn}
\end{figure}

\section{Qualitative Results on Other Image Domains}

\begin{figure}[h]
    \centering
    \includegraphics[width=\linewidth]{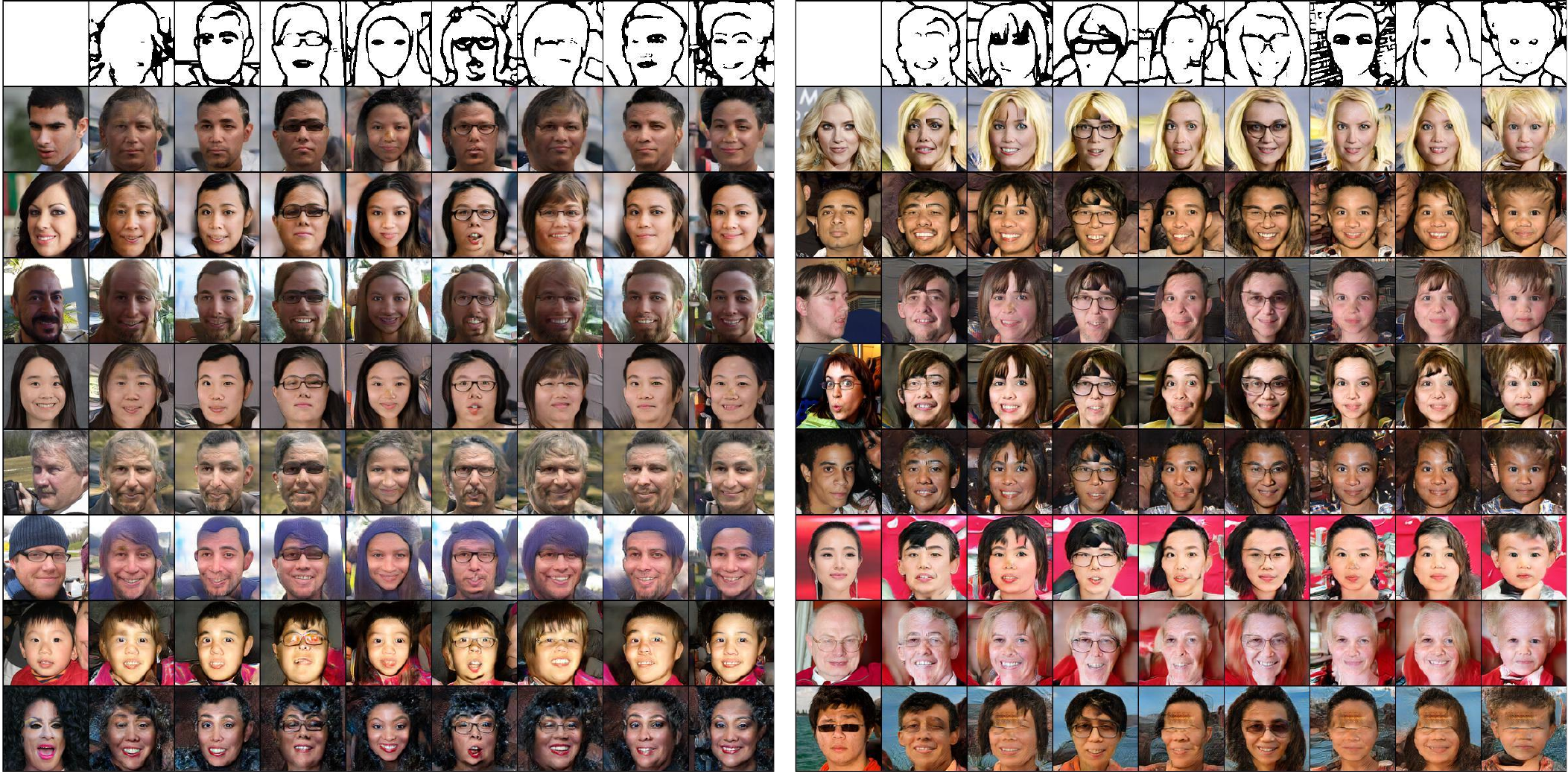}
    \caption{Qualitative results from our model trained on human face. Note that how the glasses in the sketches are successfully drawn on the generated images even when there is no glasses in the style image; and how the moustache is correctly removed when there is moustache in the style images but is not indicated in the sketches. }
    \label{fig:face_1}
\end{figure}

\begin{figure}[h]
    \centering
    \includegraphics[width=\linewidth]{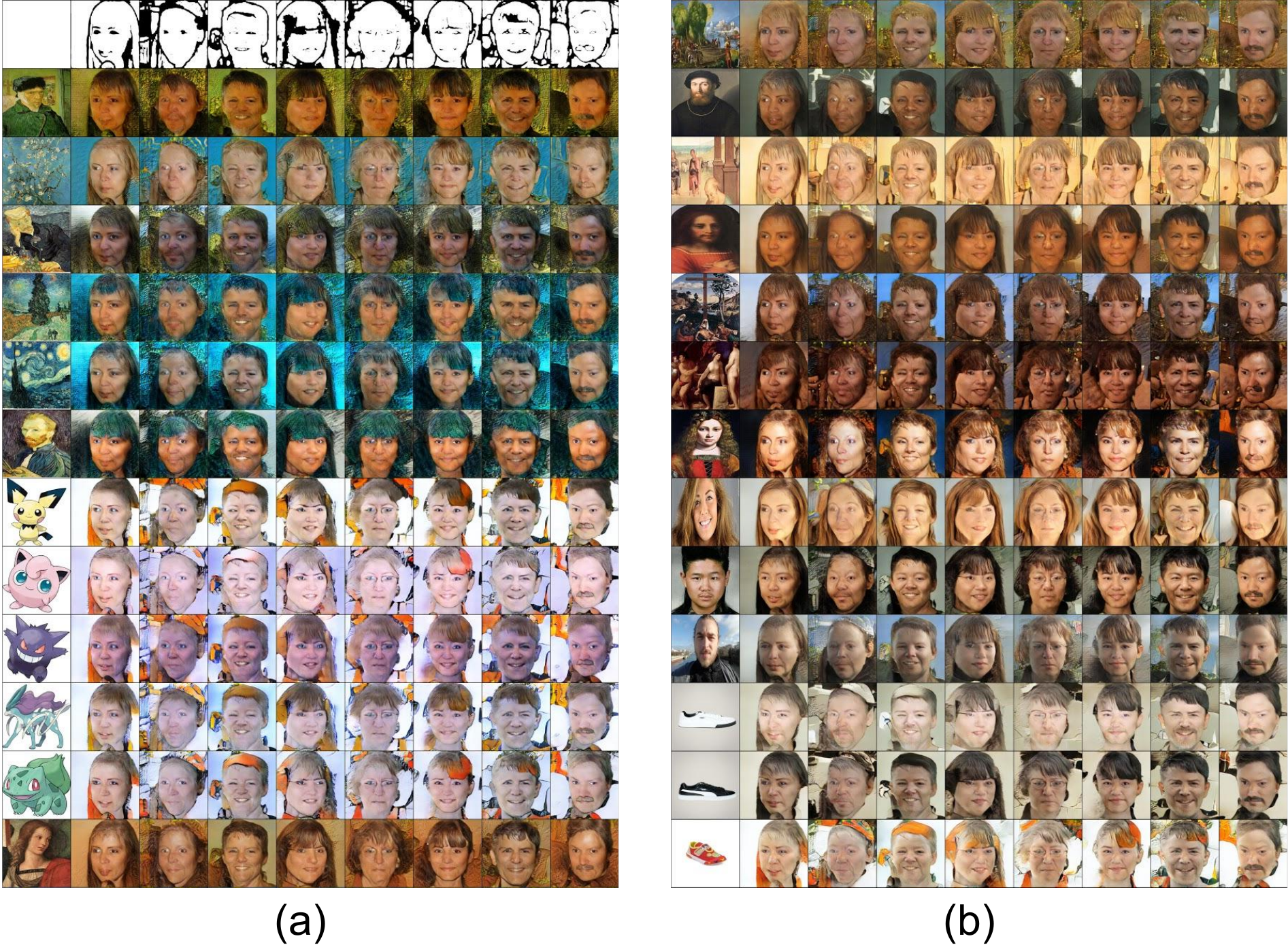}
    \caption{Qualitative results from our model trained on human face. We use random images from other domains as the style image for our model, including Pokemon, paintings, and shoes, which are not seen by the model during training. Note how the model still able to draw the essential face even when the style images do not have a face at all, showing the ability of the model learning semantics from the training data. It also shows that our model has a firm content faithfulness to the sketch. }
    \label{fig:face_2}
\end{figure}

Even though focused on art, our model is generic and can be applied to other sketch-to-image tasks. Below we show the results of our model trained on apparel and human face data (the apparel dataset is from kaggle: \url{https://www.kaggle.com/paramaggarwal/fashion-product-images-dataset}, and the human face dataset is FFHQ: \url{https://github.com/NVlabs/ffhq-dataset}). Note that since these datasets do not have the artistic style variances that we are interested in, we do not think the power of the proposed modules, especially FMT, can be adequately reflected. And we do not use the image gradient matching loss because there is no texture patterns that we want the model learn from these datasets. However, our model does show the state-of-the-art performance in general sketch-to-image tasks. Most importantly, it shows the evidence that the model learns semantics from the training corpus, as pointed out in Figure~\ref{fig:face_1} and Figure~\ref{fig:face_2}.   

\begin{figure}[h]
    \centering
    \includegraphics[width=\linewidth]{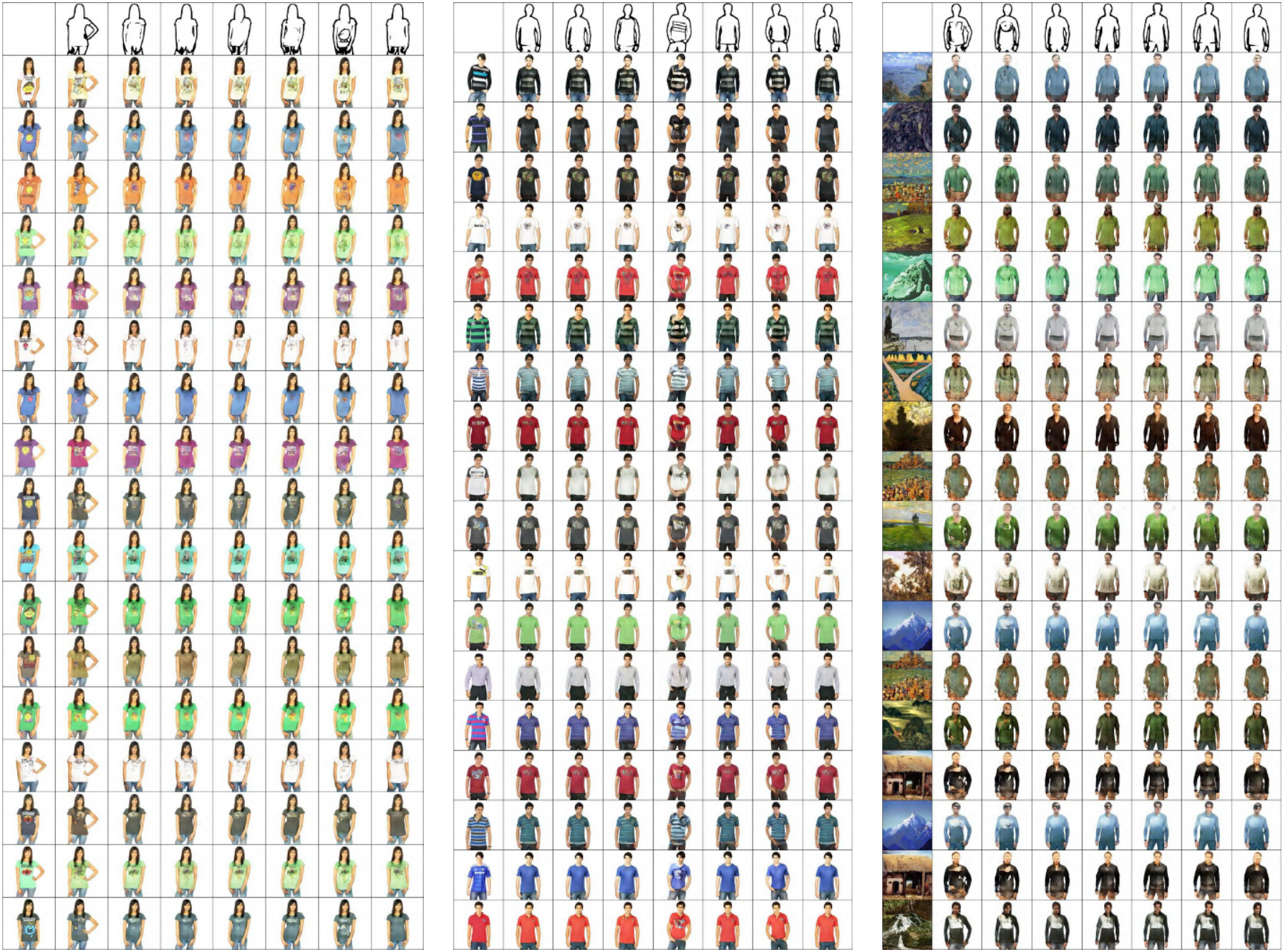}
    \caption{Qualitative results from our model trained on apparel. For the three sets of images, the first column is the style images, and the first row is the sketches. Individually, for the right-most set, we input random art images as the style image which the model never saw before. And the model is still able to get the correct shapes and colors.   }
    \label{fig:fashion}
\end{figure}

\section{Synthesis from multiple style images}
While our model is trained to take one sketch and one style image as input, there are many ways to take advantage of the model and syntheses artful image from sketches. And we believe those applications indicate the potential of our model in the field of creative art creation.

\begin{figure}[h]
    \centering
    \includegraphics[width=1\linewidth]{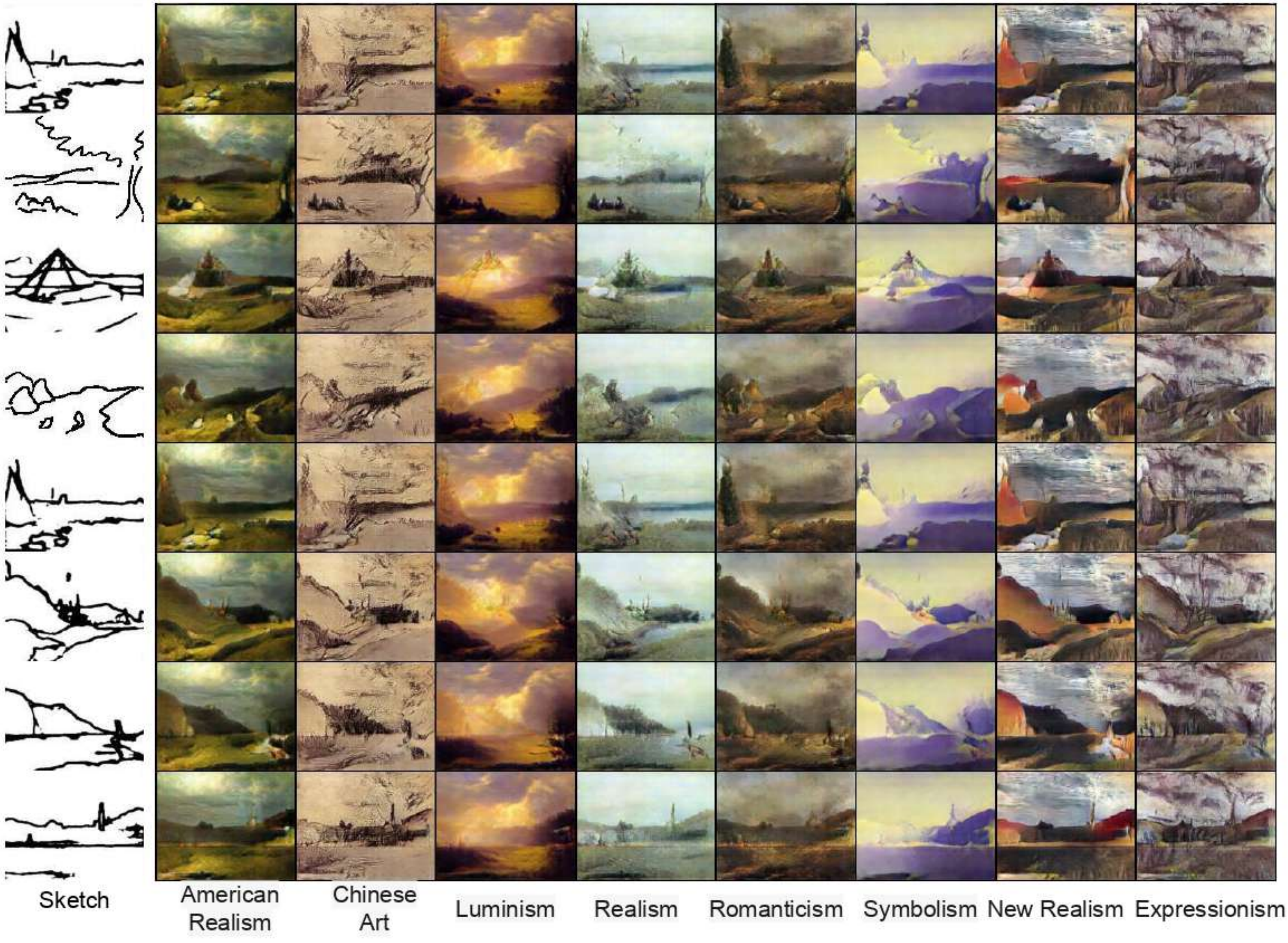}
    \caption{\textbf{Synthesizing by averaging 8 images from the same style:} We take 8 paintings from the same style, and average their extracted features from the feature extractor $E$, then use the result features for the image generation process. }
    \label{fig:avg_style}
\end{figure}

\begin{figure}[h]
    \centering
    \includegraphics[width=1\linewidth]{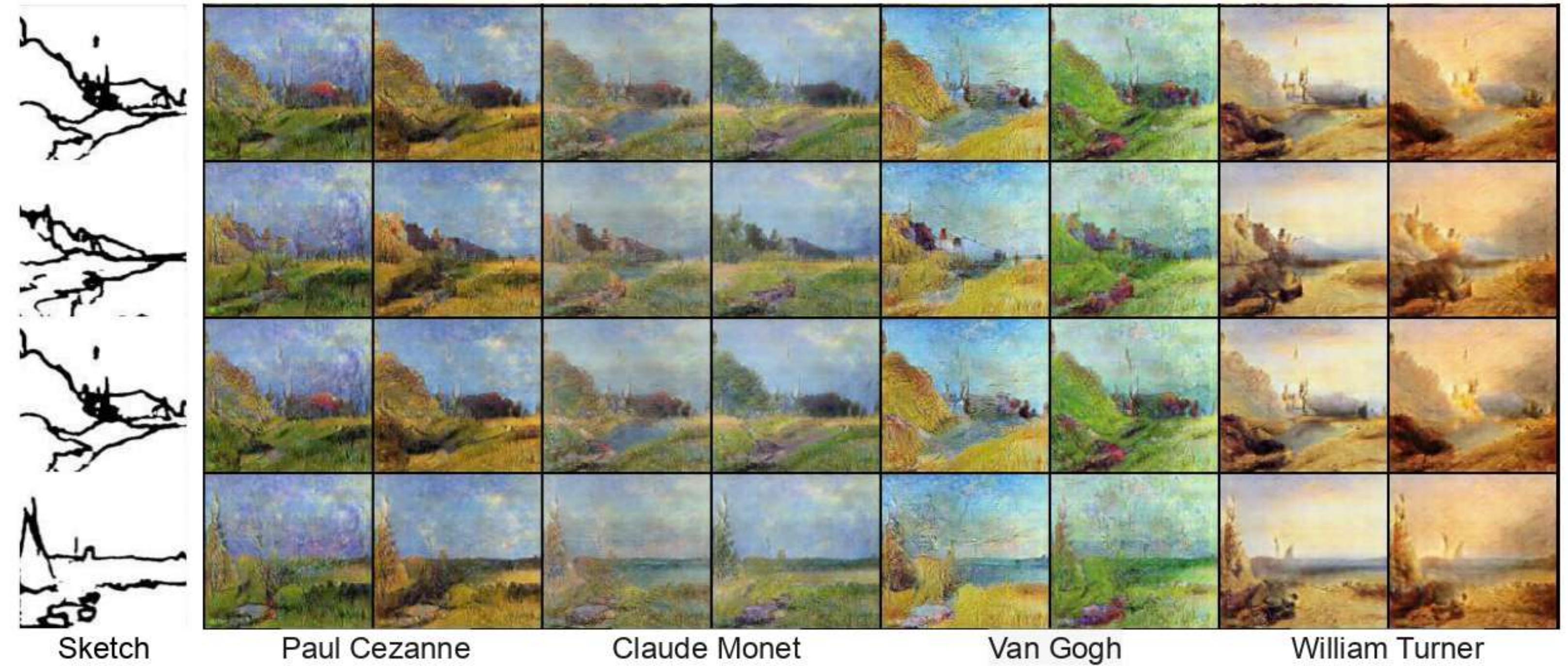}
    \caption{\textbf{Synthesizing by averaging 8 images from the same artists:} We do the same operation as in Figure~\ref{fig:avg_style} but for the same artists rather than same styles.}
    \label{fig:avg_artist}
\end{figure}

Taking a set of style images, we can let $E$ extract their features, and manipulate the set of features such as averaging them, or combine them with a given ratio, to get the summarized features. Then feed the summarized features to $G$ for the generation. Examples can be found in Figure~\ref{fig:avg_style} and Figure~\ref{fig:avg_artist}. They show that our model is able to produce high quality generation and manipulate the mixed style patterns well, thus generate meaningful images. 

Even-though trained on pair sketch/image, our model works consistently with human hand-draw sketches. Actually, we find that the extracted sketch from paintings are fairly similar to simple lines that human can draw. The generations from human-drawn sketches can be found in Figure~\ref{fig:hand_draw_1} and Figure~\ref{fig:hand_draw_2}.

\section{Discussions and Limitations}
\begin{figure}[h]
\centering
  \includegraphics[width=0.9\linewidth]{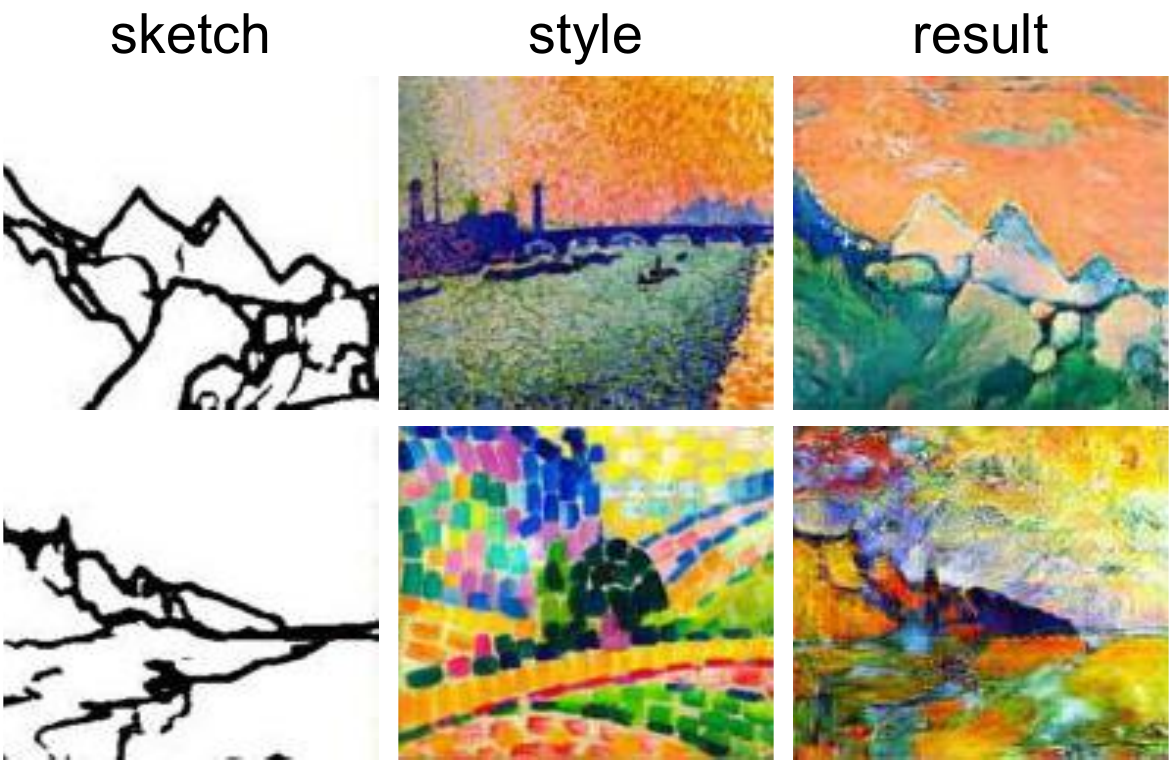}
  \caption{{\bf Limitations:} While our model gains marked progress in synthesizing artistic images from sketches, some style patterns are hard to be captured and well-represented by the current model, e.g., the pointillism style and some styles with large patches of colors. Our model has a relatively inconsistent performance on pointillism style. In row 1 where $I_{style}$ has an intense pointy texture over the whole composition, while $I_g$ is trying to imitate the pointy technique around the sky area, the result shows more of flatness across the whole image. Style with large patches of color is the one on which our model has a generally unsatisfying performance. As shown in row 2, $I_g$ can hardly reproduce the neatly arranged color patches in $I_{style}$, even though it achieves the correct color palette. We believe some dedicated style recognition and generation methods can be developed for these two styles.}
  \label{fig:discussion-1}
\end{figure}

\begin{figure}
\centering
  \includegraphics[width=1\linewidth]{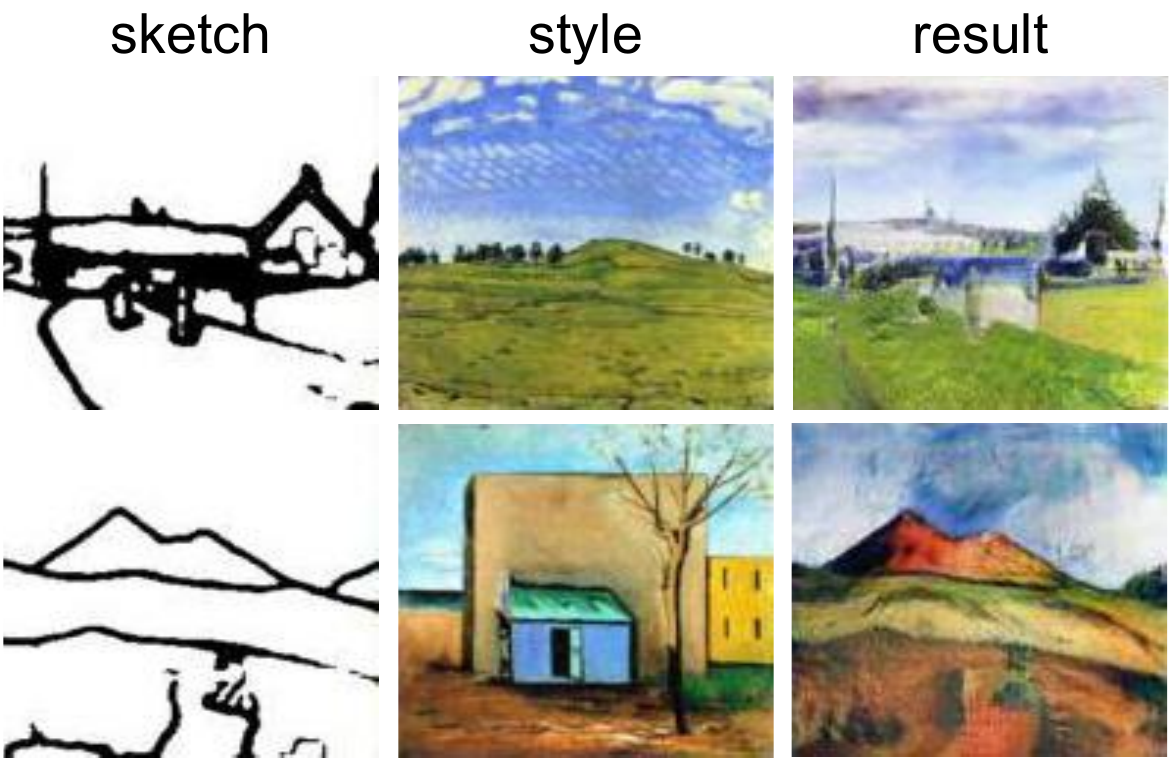}
  \caption{{\bf Effect of reference image vs. corpus:} During the training of the model, we assume that the model is able to learn some global knowledge from the whole corpus in synthesizing the images. This may contain some semantic features and more abstract style patterns. For example, in row 1, there is a house in $I_{sketch}$ but $I_{style}$ is a grassland with no color indication for the house. However, the model is still able to properly colorize the house with black roof and window. Similarly in row 2, despite that there is no mountain in $I_{style}$, the model surprisingly generates the mountain with a red color tune which is not appeared from the referential style image. Learning from the corpus can be a good thing for providing extra style cues apart from $I_{style}$, however, it may also cause conflicts against $I_{style}$, such as inaccurate coloring and excess shapes. It is worth study on how to balance the representation the model learns from the whole corpus and from the referential style image, and hoe to take advantage of the knowledge from the corpus for better generation. }
  \label{fig:discussion-2}
\end{figure}

Our discussions on the sketch to art task and some limitations in current model can be found in Figure~\ref{fig:discussion-1}, Figure~\ref{fig:discussion-2}.

\end{document}